\documentclass[11pt, a4paper, logo]{googledeepmind}

\pdftrailerid{redacted}

\makeatletter
\renewcommand\bibentry[1]{\nocite{#1}{\frenchspacing\@nameuse{BR@r@#1\@extra@b@citeb}}}
\makeatother

\usepackage{parskip}
\usepackage{amsmath, amsfonts, bm, dsfont}
\usepackage{kantlipsum, lipsum}
\usepackage{dsfont}

\usepackage[authoryear, sort&compress, round]{natbib}

\graphicspath{{figures/}}

\usepackage{amsmath,amsfonts,bm}

\def\eqref#1{Equation~(\ref{#1})}

\def\1{\bm{1}}

\DeclareMathAlphabet{\mathsfit}{\encodingdefault}{\sfdefault}{m}{sl}
\SetMathAlphabet{\mathsfit}{bold}{\encodingdefault}{\sfdefault}{bx}{n}

\DeclareMathOperator*{\argmax}{arg\,max}

\usepackage{hyperref}
\usepackage{url}
\usepackage{amsthm}
\usepackage{subfigure}
\usepackage{graphicx}

\usepackage{bbm}

\usepackage{enumitem}
\setlist[itemize]{leftmargin=0.5cm}

\newcommand{\beq}{\begin{equation}}
\newcommand{\eeq}{\end{equation}}

\newcommand{\beqa}{\begin{eqnarray}}
\newcommand{\eeqa}{\end{eqnarray}}

\newcommand{\beqan}{\begin{eqnarray*}}
\newcommand{\eeqan}{\end{eqnarray*}}

\renewcommand{\1}[1]{\mathbb{1}\{#1\}}

\newlength{\minipagewidth}
\newtheorem{hypothesis}{Hypothesis}

\newtheorem{predefinition}{Definition}

\newtheorem{theorem}{Theorem}

\renewcommand{\phi}{\varphi}

\usepackage[dvipsnames]{xcolor}

\title{Understanding the performance gap between online and offline alignment algorithms}

\keywords{Reinforcement learning from human feedback, Alignment, Offline learning, Large language models}

\reportnumber{}

\renewcommand{\today}{}

\author[1]{Yunhao Tang}
\author[1]{Daniel Guo}
\author[1]{Zeyu Zheng}
\author[1]{Daniele Calandriello}
\author[1]{Yuan Cao}
\author[1]{Eugene Tarassov}
\author[1]{R\'emi Munos}
\author[1]{Bernardo \'Avila Pires}
\author[1]{Michal Valko}
\author[1]{Yong Cheng}
\author[1]{Will Dabney}

\affil[1]{Google DeepMind}

\begin{abstract}

Reinforcement learning from human feedback (RLHF) is the canonical framework for large language model alignment. However, rising popularity in offline alignment algorithms challenge the need for on-policy sampling in RLHF. Within the context of reward over-optimization, we start with an opening set of experiments that demonstrate the clear advantage of online methods over offline methods. This prompts us to investigate the causes to the performance discrepancy through a series of carefully designed experimental ablations. We show empirically that hypotheses such as offline data coverage and data quality by itself cannot 
convincingly explain the performance difference. We also find that while offline algorithms train policy to become good at pairwise classification, it is worse at generations; in the meantime the policies trained by online algorithms are good at generations while worse at pairwise classification. This hints at a unique interplay between discriminative and generative capabilities, which is greatly impacted by the sampling process. Lastly, we  observe that the performance discrepancy persists for both contrastive and non-contrastive loss functions, and appears not to be addressed by simply scaling up policy networks. Taken together, our study sheds light on the pivotal role of on-policy sampling in AI alignment, and hints at certain fundamental challenges of offline alignment algorithms.
\end{abstract}

\begin{document}

\maketitle

\newcommand{\fix}{\marginpar{FIX}}
\newcommand{\new}{\marginpar{NEW}}

\section{Introduction}

Following the impact of large language models \citep{openai2023gpt,google2023gemini,anthropic2024claude,meta2023llama,gemma2024gemma,mistral2023mistral},
reinforcement learning from human feedback (RLHF) is now the canonical framework for AI alignment to further improve supervised fine-tuned (SFT) models  \citep{christiano2017deep,ouyang2022training}. However, recent advances in offline methods (e.g., direct preference optimization  DPO \citep{rafailov2023direct} and its variants \citep{azar2023general,zhao2023slic}, algorithms which directly align LLMs from offline dataset \emph{without} active online interactions, have proved empirically efficient in practice (e.g., see \citealp{jiang2023mistral,tunstall2023zephyr,bai2023qwen}). This raises a key question: 

\begin{center}
\emph{Is online RL necessary for AI alignment?}
\end{center}

It is of both empirical and conceptual interest to provide evidence to either answer to the above question, one way or another. From an empirical perspective, offline algorithms are much simpler and cheaper to implement than canonical online RLHF (which consists of preference modeling and sampling from the models). Hence, gathering evidence about the sufficiency of offline algorithms will pave a simpler path to AI alignment. Meanwhile, gathering evidence about the advantage of the canonical online RLHF will provide insights into the fundamental role of online interactions, and shed light on some key challenges to offline alignment.

\paragraph{Comparing online vs. offline algorithms.} Various implementation and algorithmic differences present challenges to compare online and offline methods in a fair way. For example, online algorithms tend to be more computationally intensive than offline algorithms, due to sampling and training an extra reward model. As a result, it is important to calibrate for a certain measure of \emph{budget} spent by different algorithms when measuring performance. In this work we do not prioritize compute as a main factor during comparison, and instead adopt the setting from \citet{gao2023scaling} which uses the KL divergence between the RLHF policy and reference SFT policy as a measure of budget. Across different algorithms and hyper-parameter settings, KL divergence  measures how far the RLHF policy drifts from the SFT policy in a unified way, and allows for comparing algorithms in a calibrated way.

\begin{figure*}
    \centering
    \includegraphics[width=0.86\textwidth]{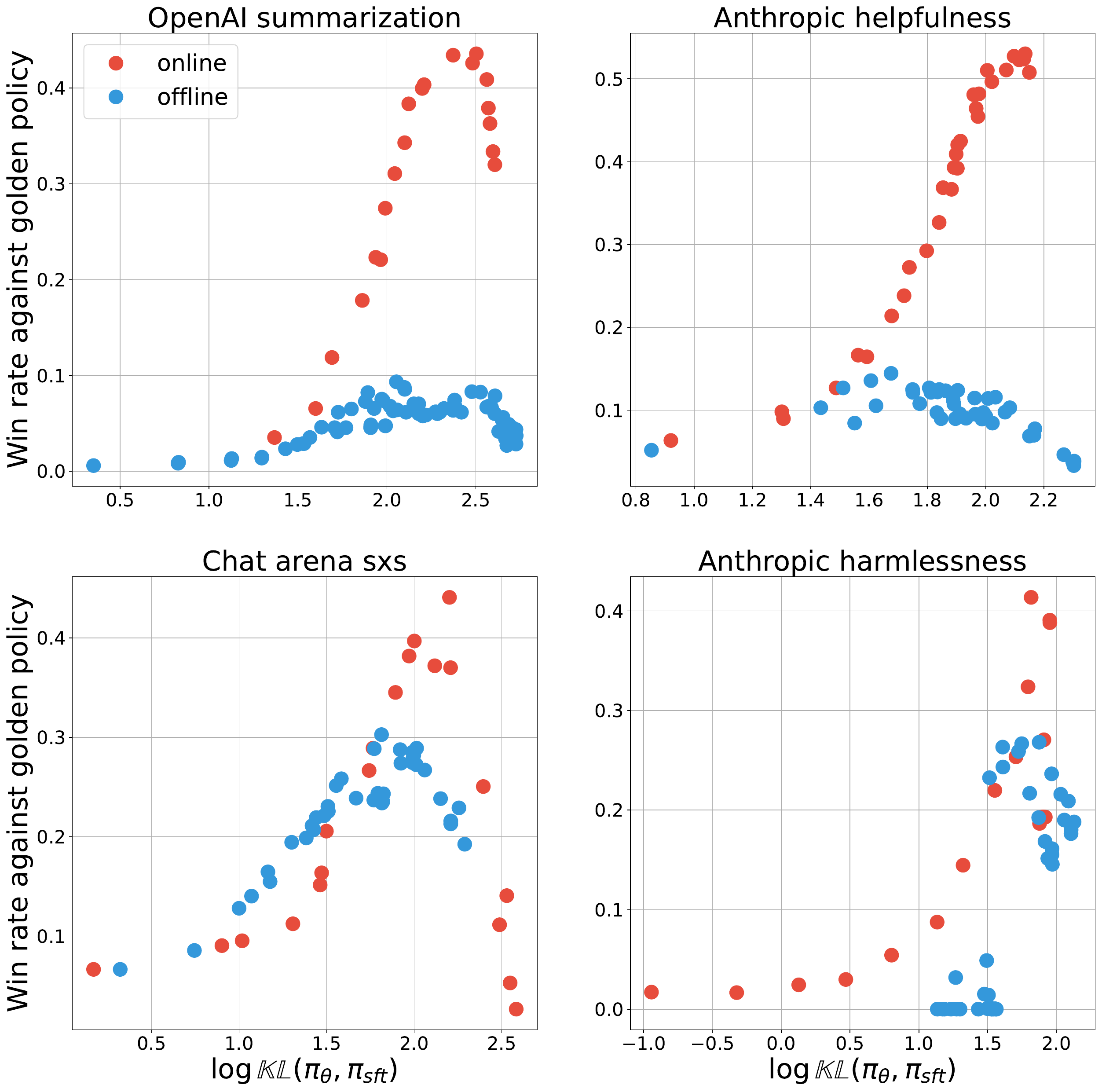}
    \caption{The trade-off between KL divergence and policy performance, for both online and offline algorithms across four open source datasets: OpenAI summarization, Anthropic helpfulness, Chat arena side by side and Anthropic harmlessness. Each data point represents the evaluation of a policy obtained during different stages of training, based on a particular set of hyper-parameters. The performance is measured as the side by side win rate of the learned policy against a golden reference policy, judged by golden preference model. As the KL divergence increases, the policy performance first increases and then decreases, consistent with \emph{Goodhart's law}. Meanwhile, online algorithms seem to generally achieve a better trade-off than offline algorithms: for a fixed budget on the KL divergence, online has generally better performance than offline. Online algorithms also achieve higher peak performance than offline algorithms across the board. See more experimental details in Section~\ref{sec:exp}.}
    \label{fig:online-offline}
\end{figure*}

\paragraph{Summary of results.}  We start with the over-optimization behavior of online vs. offline algorithms (Figure~\ref{fig:online-offline}), as predicted by extrapolations of Goodhart's law \citep{goodhart1984problems} to AI alignment \citep{amodei2016concrete}. In a controlled setup akin to \citet{gao2023scaling}, we show that on a suite of open source datasets, online algorithms generally outperform offline algorithms at the same optimization budget of KL divergence against the SFT policy (Figure~\ref{fig:online-offline}). Across different levels of KL divergence, online algorithms also generally obtain much higher peak performance than offline algorithms. In summary, online algorithms are a pareto improvement over offline algorithms, which sets the stage for the ensuing investigation.

To better understand the source of discrepancy between online and offline algorithms, we structure our investigation in the form of \emph{hypothesis testing}. That is, we start with a set of plausible hypotheses that seeking to explain for the performance gap (Section~\ref{sec:hypothesis}), and then carry out carefully controlled experiments to check if the hypotheses are valid (Section~\ref{sec:evidence}). We organize the investigation along a few dimensions and the main results are summarized as follows:
\begin{itemize}
    \item \textbf{Data}. We investigate a few hypotheses regarding the nature of offline dataset.  This includes the hypothesis that offline dataset has smaller coverage than online generated dataset (Figure~\ref{fig:online-tandem-offline}) and the hypothesis that offline algorithms are more sensitive to offline dataset with sub-optimal absolute response quality (Figure~\ref{fig:offline-dataset-golden}). Though these hypotheses are technically sensible, we empirically find that they fail to convincingly explain the performance gap. Through dataset ablations, we find that one working recipe to improve offline optimization is to generate data with distributional proximity to the starting RLHF policy (which in our case happens to be the SFT policy) (Figure~\ref{fig:offline-dataset}), which essentially mimics the initial stage of an online algorithm.
    \item \textbf{Optimization property}. We find an intriguing interplay between discriminative and generative abilities: despite being better at classification than online policy, the offline policy generates worse responses (Figure~\ref{fig:classification-rm}, Figure~\ref{fig:classification-winrate}, Figure~\ref{fig:classification}). Both across and within the same class of experiments, there appears little correlation between classification and generative performance  (Figure~\ref{fig:classification-winrate}). While both optimizing for a discriminative objective, offline sampling improves classification accuracy on a static dataset while on-policy sampling improves generative quality by constantly shifting the sampling distribution. We collect evidence showing that the improvement in generative performance is much less direct for offline than for online.
    \item \textbf{Loss function and scaling}.  To ensure that the results hold more generally, we study both contrastive and non-contrastive loss functions for RLHF (Figure~\ref{fig:bo2}). The online vs. offline performance gap seems to persist in general, though the underlying causes to such discrepancy might be likely algorithm dependent. We also study how the performance gap changes as the policy network scales up (Figure~\ref{fig:scaling}, Figure~\ref{fig:best-scaling}). The persistence of the performance discrepancy implies that the sampling issue is likely not to be addressed by simply scaling up models.
\end{itemize}

While the empirical evidence alludes to the fundamental importance of on-policy sampling for model alignment, we hope our results help unveil the empirical inner workings of offline alignment algorithms, and shed light on the performance difference. Taken together, these findings present interesting insights and challenges to RLHF practitioners, and pave the way to more efficient AI alignment practice.

\paragraph{Structure of the work.} The paper is organized as follows:
\begin{itemize}
    \item \textbf{Set the stage}. Section~\ref{sec:comparison} presents the over-optimization phenomenon for both online and offline algorithms which sets the stage of the ensuing study.
    \item \textbf{Hypotheses}. To initiate the \emph{hypothesis testing},  Section~\ref{sec:hypothesis} presents multiple hypotheses  seeking to explain the performance gap between online and offline algorithms. These hypotheses cover the aforementioned dimensions such as data, optimization property, loss function and scaling. They are designed to be verifiable to certain degree, and capture some basic instincts to the problem.
    \item \textbf{Experimental setup}. Section~\ref{sec:exp} dives into the detailed experimental setups: the controlled setup to study over-optimization and the loss function that instantiates online vs. offline algorithms.
    \item \textbf{Analysis}.  Section~\ref{sec:evidence} discusses experimental designs and evidence for or against the hypotheses proposed before. This provides an assessment of how convincing the hypotheses are, shedding some light on the nature of the online vs. offline performance gap. The analysis is followed by Section~\ref{sec:improve} that introduces a final set of dataset ablation that leads to the finding that making dataset more on-policy is a robust way to improve offline performance.
\end{itemize}

\paragraph{Isn't it obvious that online is better than offline according to existing RL literature?} The performance differences between online and offline RL algorithms have been well-establiashed in the literature \citep{levine2020offline,fujimoto2019off,ostrovski2021difficulty}, so what is surprising here? Most importantly, note that online RLHF algorithms rely on a learned reward model trained from the \emph{same} pairwise preference dataset as the offline RLHF algorithm. This differs fundamentally from regular RL settings where one assumed access to ground truth reward in online interactions, in which case online RL has a clear advantage. Assuming RLHF is bottlenecked by the reward signal \citep{gao2023scaling}, it is not clear if the online vs. offline gap would be as prominent here. 

On a more technical note, many RLHF algorithms adopt the contextual bandit formulation and apply regularization against a reference policy (e.g., all the offline methods \citep{rafailov2023direct,azar2023general,zhao2023slic,tang2024generalized}). Such algorithmic details deviate RLHF practices from regular RL settings, which are likely to impact the severity of the off-policy learning problem.

\paragraph{Limitations.} Our study has a few limitations. We have focused on open source datasets and a controlled RLHF setting, which might confound the complexity of the reward modeling task. An interesting direction is to study how the online vs. offline comparison changes based on the nature of the dataset. We also do not experiment with state-of-the-art pre-trained and post-trained models, which might impact the conclusion to some level.
Nevertheless, we expect some of the results and intuitions obtained through our controlled experiments to be more transferable in general.

\section{Comparing online and offline performance under Goodhart's law} \label{sec:comparison}

We start with an empirical comparison of online and offline alignment methods on a suite of open source datasets. Throughout this work, we focus on the IPO loss which can instantiate both online or offline algorithms depending on the source of samples \citep{azar2023general,calandriello2024human}. Similar experimental investigations can be carried out for DPO \citep{rafailov2023direct} or more general contrastive algorithms \citep{tang2024generalized}, though we expect our results to reasonably generalize due to similar empirical patterns of other algorithms \citep{tang2024generalized}.

We provide a brief background on the  IPO loss. In general, we assume a prompt distribution $x\sim p$ and a sampling distribution from which to draw responses $y,y'\sim\mu(\cdot|x)$. The responses are then ranked either by a proxy preference model or by human rater into a winning and a losing response $(y,y')\rightarrow (y_w,y_l)$. The algorithm minimizes the loss
\begin{align}
   \min_\theta \mathbb{E}_{x\sim p,(y_w,y_l)\sim\mu}\left[\left(\log\frac{\pi_\theta(y_w|x)}{\pi_\text{sft}(y_w|x)} -\log\frac{\pi_\theta(y_l|x)}{\pi_\text{sft}(y_l|x)} - \frac{\beta}{2}\right)^2\right],\label{eq:loss}
\end{align}
with respect to the policy parameter $\theta$. Here, $\pi_\text{sft}$ denotes the SFT policy, used as the reference policy during training.  The constant scalar $\beta>0$ is a hyper-parameter that controls the strength of regularization, the bigger the value of $\beta$ the stronger the regularization. 
The intuition of the algorithm is that by minimizing the loss, $\pi_\theta$ places more weights on the winning response $y_w$ relative to the losing response $y_l$. We provide more detailed discussions in Appendix~\ref{appendix:algorithm}. 

Online and offline algorithms differ in the implementations of sampling distributions $(p,\mu)$: while both draw prompts uniformly from a fixed dataset which defines the distribution $p$, the offline algorithm draws responses from a fixed dataset which also implicitly defines $\mu$ \citep{rafailov2023direct,azar2023general}; meanwhile, the online algorithm samples the responses on-policy $\mu=\pi_\theta$ \citep{calandriello2024human,guo2024direct}. Except for the difference in the sampling process, online and offline algorithms are identical in their loss functions and share the same set of hyper-parameters. This allows us to conduct experiments in controlled settings for fair comparison.

We choose to study contrastive losses rather than a policy gradient loss such as PPO \citep{schulman2017proximal,openai2023gpt}. The motivations for such a choice are rather technical: it is not clear how to optimize the PPO loss offline since the algorithm by design uses on-policy samples; in fact, DPO was intially derived as an offline equivalent of such policy gradient based algorithms \citep{rafailov2023direct}. By studying IPO, we also avoid the complications of learning value functions. Despite the technical differences, we expect some insights here to transfer to other RLHF algorithms.

\subsection{Online achieves better budget and performance trade-off than offline}
As mentioned in the introduction, to account for the effective budget spent by different algorithms, we adopt the synthetic setup introduced in \citet{gao2023scaling} to study the trade-off between the KL divergence $\mathbb{KL}\left(\pi_\theta,\pi_\text{sft}\right)$ and the policy performance. The KL divergence measures how far the optimized policy $\pi_\theta$ drifts away from the supervised fine-tuned (SFT) policy $\pi_\text{sft}$ during training, and can be understood as an optimization budget consumed by the algorithm. 

Since both online and offline algorithms optimize for a proxy objective, either in the form of learned reward model or finite offline dataset, as the optimization progresses both algorithms will over-optimize and degrade their performance on the ground truth performance,  which we measure as the win rate against a golden policy baseline using a golden preference model (details to be introduced later). Such a general phenomenon of \emph{over-optimization} can captured by a slight extrapolation of Goodhart's law, which states that \emph{when a measure becomes a target, it ceases to be a good measure} \citep{goodhart1984problems}.

In Figure~\ref{fig:online-offline} we present the trade-off between KL divergence and policy performance for both online and offline algorithms, across four different open source datasets. Each data point in the plot shows the evaluation of a policy at a particular checkpoint during training, for a particular set of hyper-parameters. For the online algorithm, we did not extensively tune the hyper-parameter and used a fixed set of hyper-parameter throughout, while for the offline algorithm we pooled together results for sweeps over hyper-parameters. A few observations are in order:
\begin{itemize}
    \item \textbf{Over-optimization under Goodhart's law}. For both online and offline algorithms, the performance first increases and then decreases as a function of the KL divergence. The decrease is due to the over-optimization effect, captured by the Goodhart's law as discussed above \citep{amodei2016concrete,gao2023scaling}.
    \item \textbf{Online uses KL divergence budget more efficiently than offline}. Online algorithms seem to generally achieve a better trade-off compared to offline algorithms. Concretely, with the same budget on the KL divergence, online algorithms obtain generally better performance than offline algorithms. Across different KL levels, online algorithms' peak performance is higher than that of the offline algorithms across all tasks. The improvement is more profound for the OpenAI summarization and Anthropic helpfulness task, while for the other two the peak difference is smaller.
\end{itemize}

Note that the performance of the over-optimization curves in Figure~\ref{fig:online-offline} is measured as the win rate in a side-by-side comparison against a fixed baseline (details in Section~\ref{sec:exp}), and hence not an apple-to-apple comparison to the results from \citet{gao2023scaling} where they use rewards as the performance measure. Nevertheless, taking together results across all four datasets, we can conclude a statistically significant performance gap between online and offline algorithms, the cause of which we  investigate below.

\section{Hypotheses for the performance discrepancy} \label{sec:hypothesis}

\begin{figure*}
    \centering
    \includegraphics[width=0.95\textwidth]{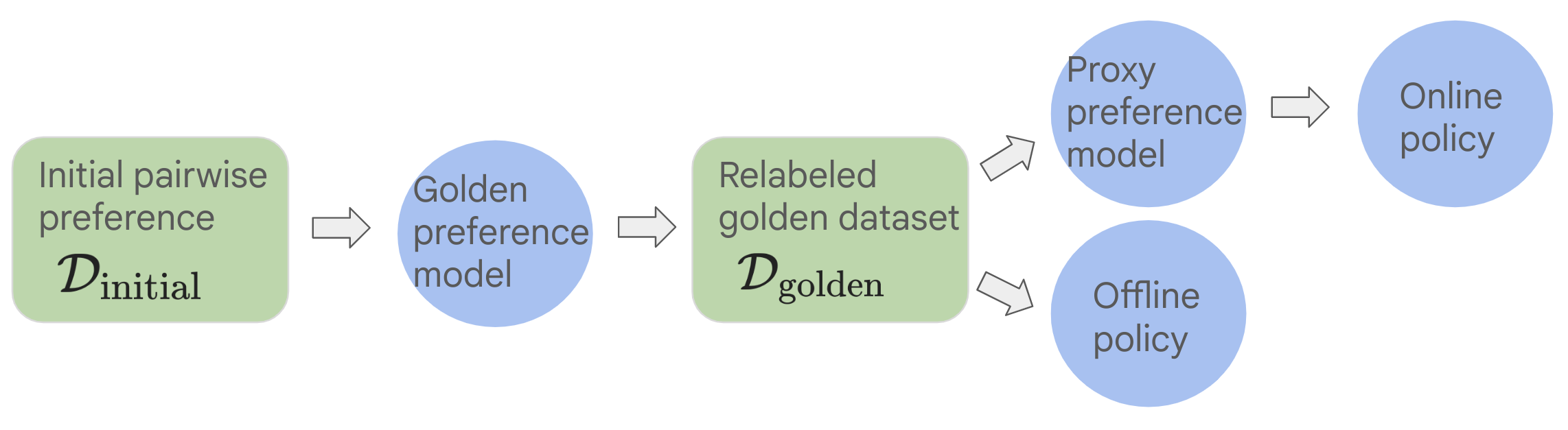}
    \caption{Controlled experimental setting in this work, adapted from \citep{gao2023scaling}. Green boxes indicate datasets and blue boxes indicate learned preference models or policies. Starting with an initial pairwise preference dataset (e.g., OpenAI summarization dataset) $\mathcal{D}_\text{initial}$, we train a XXL golden preference model that is used for relabeling the original dataset. The newly relabeled dataset $\mathcal{D}_\text{golden}$ is used for downstream policy optimization and preference model learning. The upper path depicts the online algorithms: from $\mathcal{D}_\text{golden}$, we learn a large preference model, followed by online optimization against the proxy preference model. The lower path depicts the offline preference optimization, which directly optimizes the loss function on $\mathcal{D}_\text{golden}$.}
    \label{fig:workflow}
\end{figure*}

Why are online algorithms seemingly better than offline algorithms, despite the fact that they are \emph{grounded} on the same set of pairwise preference data to start with
(i.e., the same pairwise preference dataset for training preference model for online algorithms and training policy for offline algorithms, see Figure~\ref{fig:workflow} for a visual depiction of the online vs. offline algorithm workflow)? 

Before diving into details of the experimental designs in Section~\ref{sec:exp}, we provide a few reasonable hypotheses. These hypotheses are not intended to be mathematically precise, yet seek to capture intuitive (though maybe untrue) explanations to the performance discrepancy. These hypotheses are not exhaustive, as they are designed to be relatively verifiable through ablation study.

\begin{hypothesis}
\textbf{Data coverage}.
Online algorithms are better because the data coverage is more diverse (i.e., sampled from different learner policies over time) than the offline dataset.
\end{hypothesis}

A first hypothesis is based on the intuition that online algorithms might somehow benefit from sampling a more \emph{diverse} set of responses compared to the offline dataset. Unlike offline algorithms which are constrained to a static dataset of offline responses, online algorithms have access to many more responses generated by the model. The diversity of the responses might also come from the fact that the policy keeps evolving over time, and can generate a wider coverage compared to the static offline dataset. 

\begin{hypothesis}
\textbf{Sub-optimal offline dataset}.
Offline algorithms are at a disadvantage because the initial preference dataset is generated by a sub-optimal policy. If offline algorithms are trained with responses with higher \emph{absolute} quality, the performance will be better.
\end{hypothesis}

To understand the motivation behind the hypothesis, we may interpret offline algorithms as a contrastive version of SFT: increasing the probability of winning responses $y_w$ while decreasing the probability of losing responses $y_l$. This interpretation can be deduced from the loss function that aims to increase the log ratio \citep{rafailov2023direct,azar2023general,zhao2023slic,tang2024generalized} 
\begin{align}
    \log \frac{\pi_\theta(y_w|x)}{\pi_\theta(y_l|x)} = \underbrace{\log \pi_\theta(y_w|x)}_{\text{increase winning prob}} - \underbrace{\log \pi_\theta(y_l|x)}_{\text{decrease losing prob}} \label{eq:contrastive-sft}
\end{align}
Hence, it feels that offline algorithms are inefficient if the dataset itself is generated from sub-optimal policies. Though the same argument applies to learning preference model from the very same dataset, it might be that somehow policy learning is more sensitive to dataset quality.

\begin{hypothesis}
\textbf{Better classification better performance}. Offline algorithms typically train policies as classifiers. However, as classifiers they might not be as accurate as proxy preference models (due to effectively different ways to parameterize the classification). If the accuracy improves, the performance will improve too.\end{hypothesis}

By construction of contrastive alignment algorithms, offline algorithms train policies as reward functions, or more generally classifiers of pairwise responses (see, e.g. derivation in \citep{rafailov2023direct} and Appendix~\ref{appendix:policy-as-classifier} for how to use policy as classifier). In the meantime, online algorithms take the extra step of training proxy reward or preference models as classifiers.

Hence, it might just be the case that somehow the proxy preference model obtains higher  classification accuracy compared to the offline policy , and hence provide more informative signals to the online algorithms. If this hypothesis were true, one can expect to improve the offline policy performance by improving the classification accuracy.

\begin{hypothesis}
\textbf{Non-contrastive loss function}. How much of the performance gap is attributed to the loss function being contrastive, rather than samples being offline?
\end{hypothesis}

Though the dominant class of RLHF loss functions is contrastive (partly due to the dominant formulation of pairwise comparison as the form of human feedback \citep{christiano2017deep}), one might question how much of the performance is due to the loss function being contrastive, rather than the algorithm being offline. In other words, if we switch to a non-contrastive loss function, such that the update is more akin to SFT, the online vs. offline performance difference will hopefully decrease.

\begin{hypothesis}
\textbf{Scaling policy is all you need}. Scaling policy size upwards is all you need to bridge the gap between online and offline algorithms.
\end{hypothesis}

Recent years have constantly witnessed emergent capabilities of large models when scaled up \citep{radford2018improving,wei2022emergent}. Larger models may exhibit qualitatively distinct properties from the small ones, and hence by scaling up policy networks, we might already have vanished the performance gap between online vs. offline algorithms. If this were true, sampling would not be an issue for larger models, and maybe offline alignment suffices when large model training becomes more affordable.

\section{Experimental settings}\label{sec:exp}

In this section, we introduce the experimental setup and terminologies that will facilitate later discussions. All experiments are carried out with T5X models \citep{raffel2020exploring} with the T5X data and compute framework \citep{roberts2023scaling}. We study four tasks for a good coverage of RLHF problems: OpenAI summarization  \citep{stiennon2020learning},  Anthropic helpfulness \citep{bai2022constitutional}, Chat arena side by side \citep{chiang2024chatbot}, Anthropic harmlessness \citep{bai2022constitutional}.

\subsection{Controlled setting to study KL vs. performance trade-off} \label{sec:exp-controlled}

We describe the overarching design to study the trade-off between KL divergence and policy performance (results in Figure~\ref{fig:online-offline}) in a controlled environment (see Figure~\ref{fig:workflow} for a visual depiction). The setup is similar to \citet{gao2023scaling}.

For a given task, we are provided with an initial offline dataset consisting of prompts and labeled pairs of responses $\mathcal{D}_{\text{initial}}$ (throughout, we use only the train split from such datasets since we do not study generalization to test split prompts). We follow the experimental setup from \citet{gao2023scaling} and train a \emph{golden} preference model with a XXL T5X model (11B parameters). Then we label the train split with the golden preference model and create a new dataset $\mathcal{D}_{\text{golden}}$ as the starting point of the investigation, used for downstream online and offline algorithms. By design, the golden preference model trains with the largest T5X model and is meant to emulate ground truth preferences and as a simulated alternative to human raters. We clearly distinguish the \emph{golden} reward from any \emph{proxy} preference models downstream. We specify more details about preference model training in Appendix~\ref{appendix:preference}.

Then the golden dataset is used for either training the preference model for the online algorithm, or for training policy directly for the offline algorithm. Unless otherwise mentioned such as in the scaling experiments, both the policy and proxy preference model are Large T5X models (with 770M parameters) initialized from supervised checkpoints (Appendix~\ref{appendix:sft}). 

As discussed before, we treat the KL divergence $\mathbb{KL}\left(\pi_\theta,\pi_\text{sft}\right)=\mathbb{E}_{x\sim \rho,y\sim\pi_\theta(\cdot|x)}\left[\log \frac{\pi_\text{sft}(y|x)}{\pi_\theta(y|x)}\right]$ as a proper measure of distance between learned policy and SFT policy. In practice, we estimate the KL divergence by subsampling 256 prompts from the training set, and constructing one sample-based unbiased estimate. We leave more details in Appendix~\ref{appendix:kl}.

\paragraph{Supervised fine-tuning.} As the starting point of the RLHF process, we carried out supervised fine-tuning of the policy against the aggregate of all responses from the pairwise preference datasets. This SFT stage is not aimed for improving the model quality, rather it is meant to just ensure that from the beginning of RLHF, the policy can already produce reasonable samples with a wide coverage. This also ensures that our overall experimental designs are closer to realistic scenarios. A future direction is to investigate how the conclusions change as the SFT model quality varies.
See Appendix~\ref{appendix:sft} for more technical details.

\paragraph{Evaluation.} Throughout, we evaluate the performance of any learned policy by the win rate against a fixed policy baseline over $2048$ prompts sub-sampled in a deterministic way. This fixed policy baseline is trained via the online algorithm with the golden preference model with the baseline configuration specified above. Because of the use of the golden preference model, this policy baseline has access to privileged information and is the gold standard for policy performance. The win rate is determined by the golden preference model. 

\paragraph{Hyper-parameters.}
Throughout, we adopt the baseline hyper-parameter configuration of training for 4k gradient steps, at learning rate $1\cdot 10^{-5}$ and regularization coefficient $\beta=0.1$. This is partly adapted from prior work \citep{munos2023nash,calandriello2024human} with extensive tuning on the OpenAI summarization task. We did not tune the hyper-parameters for other tasks though noticed that Anthropic harmlessness requires generally fewer steps to learn. We will vary hyper-parameter configuration for ablation study, specifically for offline algorithms.

To ablate on the hyper-parameters of offline algorithms, starting from the baseline hyper-parameter configuration, we vary the learning rate ($3\cdot 10^{-6},1\cdot 10^{-5},3\cdot 10^{-5}$), regularization coefficient ($0.1,0.5,1$) and training steps ($4k,20k$ steps) for the offline experiments. These different combinations lead to a wider span over the KL divergence during training, and allow us to assess the maximum possible performance of offline training given a KL divergence budget.

\section{Investigating the hypotheses} \label{sec:evidence}

We now present a series of ablations to verify whether the above hypotheses provide valid explanations to the online vs. offline performance gap.

\begin{figure*}
    \centering
    \includegraphics[width=0.95\textwidth]{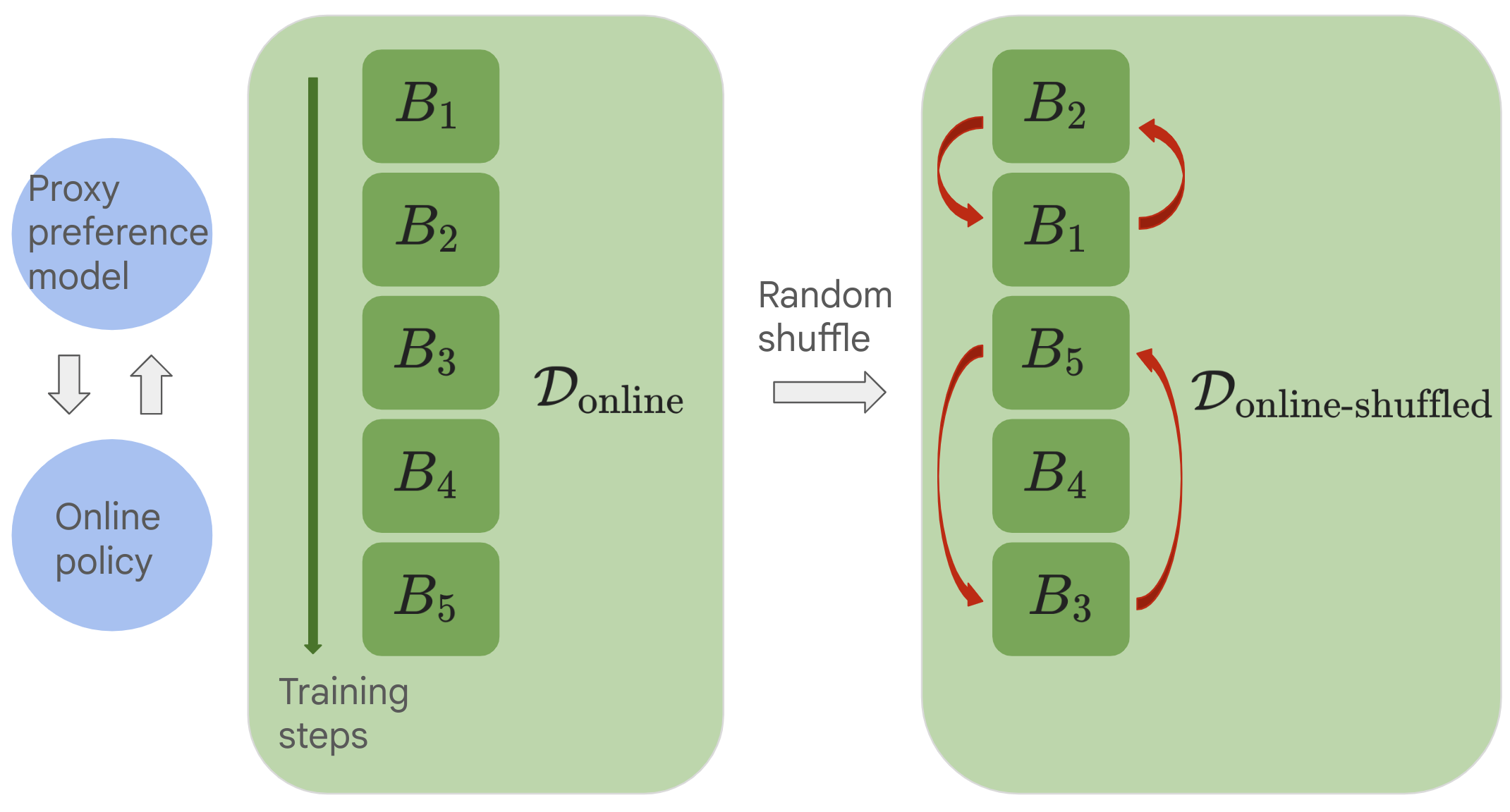}
    \caption{Illustration of online generated dataset. The online algorithms mainly consist of the interaction between the proxy preference model and online learned policies. The policy generates a sequence of data, illustrated as $B_1, B_2, B_3, B_4, B_5$, each being a tuple of a prompt, a sampled pair of responses and the preference produced by the proxy preference model. This dataset, with the online data order preserved, is denoted $\mathcal{D}_\text{online}$. The shuffled dataset $\mathcal{D}_\text{online-shuffled}$ is obtained by randomly shuffling the online dataset, in this case producing the randomized sequence $B_2, B_1, B_5, B_4, B_3$.}
    \label{fig:online-generated-dataset}
\end{figure*}

\subsection{Evidence against Hypothesis 1 (Data coverage)}

To study the effect of dataset coverage, we carefully design an experimental setting in between the online and baseline offline experiments: in a nutshell, we save all the data generated by online algorithms, and wrap it into an offline dataset for offline algorithms. This setting is heavily inspired by the study of the \emph{tandem effect} in off-policy RL \citep{ostrovski2021difficulty}.

More concretely, the  online algorithm is run with the proxy preference model on prompts from $\mathcal{D}_\text{golden}$. The online algorithm creates a stream of data $(x_t,y_t^{(w)},y_t^{(l)})_{t=1}^T$ over time, with each $(x_t,y_t^{(w)},y_t^{(l)})$ being a batch of $B=32$ prompts and responses labeled by the proxy preference model. This sequence of data forms the \emph{online generated dataset} $\mathcal{D}_\text{online}$, which by default preserves \emph{exact} ordering of the online data stream. A few comments are in order to clarify the setup:
\begin{itemize}
    \item \textbf{Online algorithm $=$ offline algorithm with $\mathcal{D}_{\text{online}}$}. It is helpful to see that by running offline algorithms on $\mathcal{D}_{\text{online}}$, we expect to recover the exact sequence of learning steps as the online algorithm. This is because both optimization processes start with the same initialization $\pi_\text{sft}$ and consume the exact same sequence of data, hence producing the same sequence of updates and parameters. This offline setup is purposely designed since general offline algorithms cannot have access to the online data order. We have validated that empirically the two procedures are indeed equivalent, up to numerical randomness due to hardware which we detail in Appendix~\ref{appendix:tpu-tandem}.
    \item \textbf{Introducing $\mathcal{D}_{\text{online-shuffled}}$: a randomly shuffled $\mathcal{D}_{\text{online}}$}. A more \emph{realistic} dataset is when the online dataset is randomly shuffled and we label such a dataset as tandem dataset $\mathcal{D}_{\text{online-shuffled}}$ (see Figure~\ref{fig:online-generated-dataset} for a graphical depiction). Such a dataset has the same data coverage (prompts, responses and preferences) as the online dataset $\mathcal{D}_\text{online}$ with the only difference that examples are presented in a random order, rather than a particular online order from the online dataset.
\end{itemize}

\begin{figure*}
    \centering
    \includegraphics[width=0.86\textwidth]{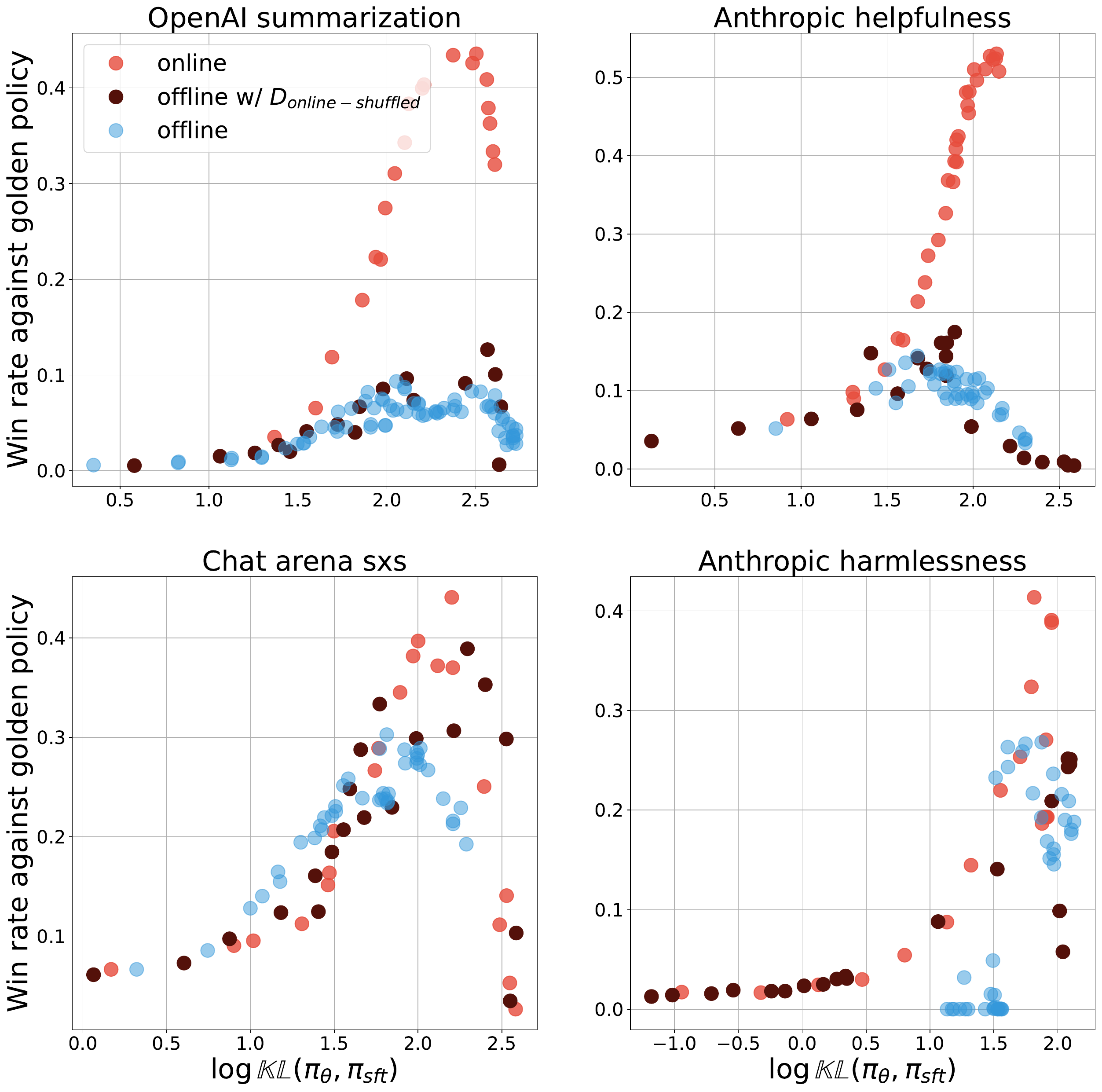}
    \caption{Evidence against Hypothesis 1 (data coverage). We save all the online generated data into an offline dataset $\mathcal{D}_{\text{online}}$ and run offline algorithms on a shuffled version of the dataset $\mathcal{D}_{\text{online-shuffled}}$. The results above compare such offline algorithms against the online and offline algorithms in Figure~\ref{fig:online-offline}. Despite the fact that such offline experiments have the same data coverage as the online experiments, there is still a significant performance gap. As a result, data coverage cannot satisfactorily explain the discrepancy between online and offline experiments.}
    \label{fig:online-tandem-offline}
\end{figure*}

We now compare the KL divergence vs. performance trade-off of running offline algorithms on $\mathcal{D}_{\text{online-offline}}$ against the other two baselines (online and offline). In Figure~\ref{fig:online-tandem-offline}, we showcase the new trade-off curves on top of the existing results in Figure~\ref{fig:online-offline}. For the results for offline with shuffled online data, we pooled together multiple experiments with different hyper-parameters (learning rate and regularization coefficient). The main observation is that, overall, there is little improvement on the offline performance due to the change from offline to the shuffled online dataset.

The results in Figure~\ref{fig:online-tandem-offline} suggests that the performance difference between online and offline algorithms cannot be explained by the difference in data coverage \emph{alone}. offline algorithms, even when augmented with the same data coverage as the online algorithm ($\mathcal{D}_{\text{online-shuffled}}$), cannot obtain the same level of performance. This alludes to the importance of the exact sampling order, obtained via on-policy sampling by a constantly evolving policy, which is in general not available for offline learning.

\paragraph{The case of Chat arena sxs.} Maybe a probable exception is the Chat arena sxs dataset, where the offline dataset with 
$\mathcal{D}_{\text{online-shuffled}}$ gets much closer to the online performance (accounting for randomness due to evaluation and training). In this particular case, one might conclude that as long as the data coverage is correct, the data ordering (i.e. on-policy sampling) is not as important. However, it is important to be mindful that arriving at the shuffled dataset $\mathcal{D}_{\text{online-shuffled}}$ in the first place is infeasible since one requires access to the whole path of data generated by an online algorithm.

\paragraph{On-policy sampling does not have to be exact.}
So far we have assume uniformly random shuffling of the entire dataset $\mathcal{D}_{\text{online}}$ to produce $\mathcal{D}_{\text{online-shuffled}}$. We can in general define a \emph{level} of shuffling, where at one extreme there is no shuffling and at another extreme we have uniform shuffling. We observe that as the amount of shuffling increases, the performance stays constant for a while before significantly deteriorating. One interpretation is that when the shuffling is small, the dataset is still largely on-policy and the learning process is robust to slight off-policyness. See Appendix~\ref{appendix:results} for more detailed results.

\begin{figure*}
    \centering
    \includegraphics[width=0.96\textwidth]{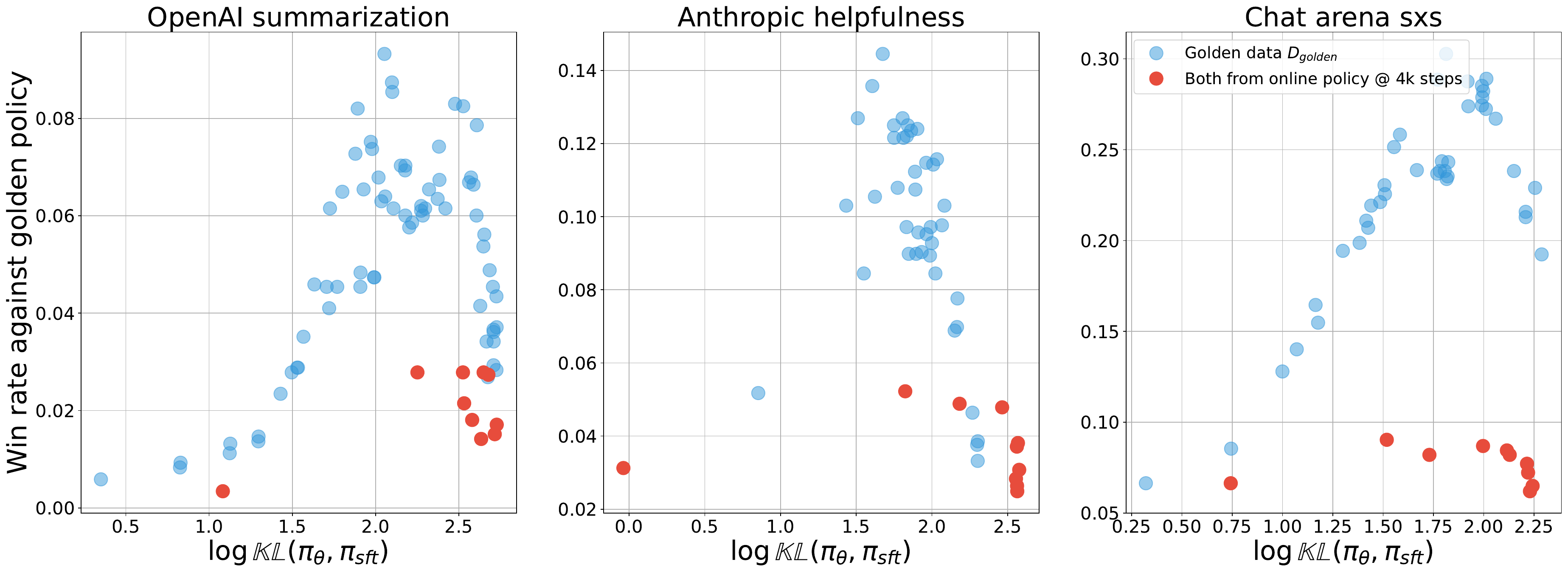}
    \caption{We vary the offline datasets and assess how the policy performance changes as a result. We compare the golden preference data $\mathcal{D}_\text{golden}$ which is the baseline, against a dataset $\mathcal{D}_\text{4k vs. 4k}$ where both sides are generated by online IPO policy trained for 4k steps. It seems that the newly generated dataset is not able to yield better performance.}
    \label{fig:offline-dataset-golden}
\end{figure*}

\subsection{Evidence against Hypothesis 2 (Sub-optimal offline dataset)}

Since we have little control over prompts and especially responses from the initial external dataset $\mathcal{D}_\text{initial}$ and the subsequent preference dataset $\mathcal{D}_\text{golden}$, it might be just that the responses were generated by sub-optimal policies, which were somehow \emph{low quality} data. This hypothesis can also provide partial explanation to the above evidence against Hypothesis 1: since the online generated dataset consists of initial responses from sub-optimal policies, through random shuffling, such sub-optimal data may be shuffled towards the end of the dataset. Learning from such data towards the end of training might be counterproductive.

If this hypothesis were true, offline algorithms would benefit from data generated by high performance policy. We hence generate an offline dataset $\mathcal{D}_\text{4k vs. 4k}$ with the following steps: with the prompt set from $\mathcal{D}_\text{golden}$, we resample paired responses from the online algorithms' final policies (online policy trained for 4k steps) which generally achieve very high win rate. Then we apply the golden preference model to relabel the generations, which produces a new dataset.

In Figure~\ref{fig:offline-dataset-golden} we show results for train offline algorithms on such a dataset. We tune the learning rate from the baseline configuration and train for 20k steps. We plot the KL divergence vs. performance trade-off under such an offline setting, in comparison to the baseline offline experiments from the pairwise dataset $\mathcal{D}_\text{golden}$. The result shows that the offline policy improves slightly compared to the SFT policy, but the win rate lingers around low values. This implies a dataset generated by high-performance policy, is not the right fix to the offline discrepancy from the online policy.

\paragraph{An empirical case that existing theory cannot properly predict.} Before moving on, we quickly note that the observation constitutes an empirical case 
where existing theories of offline algorithms fail to predict \citep{rafailov2023direct,azar2023general}. By theoretical derivations, the optimal policy to the offline problem can be understood as an one-step improvement over the dataset policy $\mu$ (the policy that generates the dataset). However, here this is not quite the case since $\mu$ has much higher performance compared to the set of possible policies obtained from the offline optimization using dataset generated by $\mu$. A conjecture is that the theoretical result fails because one typically assumes that $\mu$ has full support, which is practically invalid. See Appendix~\ref{appendix:theory-failures} for more discussions.

\begin{figure*}
    \centering
    \includegraphics[width=0.95\textwidth]{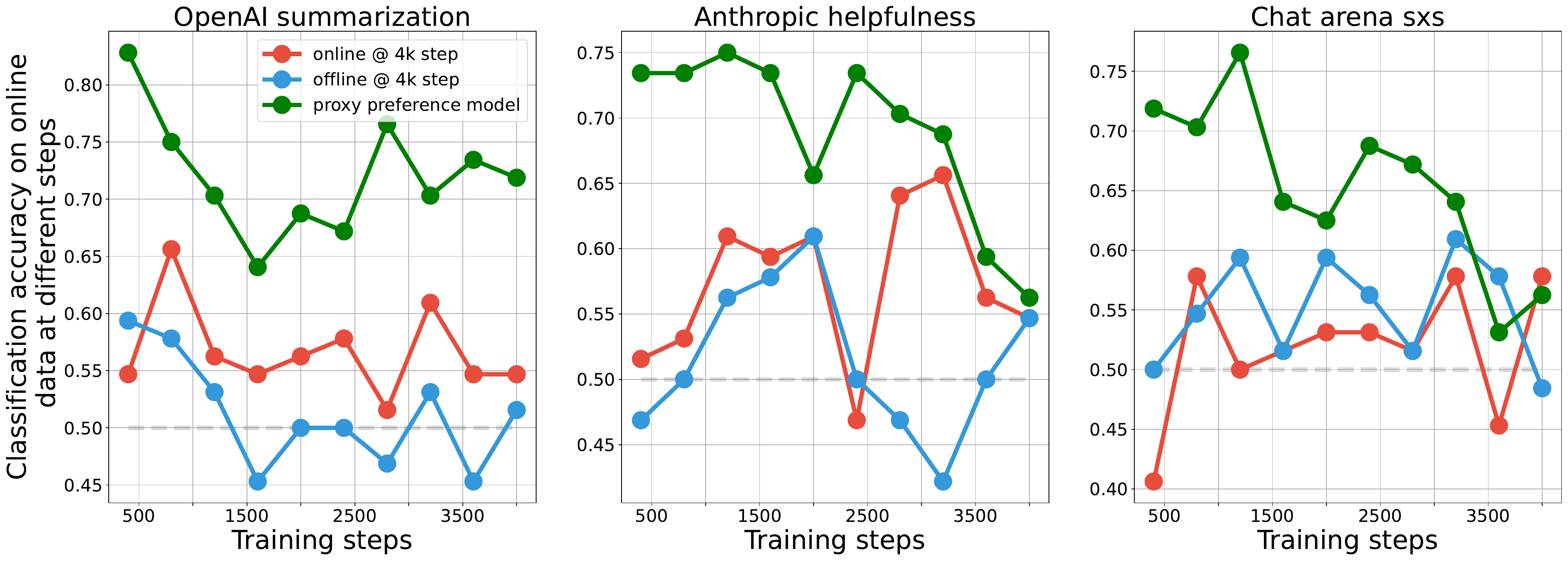}
    \caption{Classification accuracy of proxy preference models and learned policies used as classifiers by online and offline experiments (for 4k steps). The evaluation is over the sequence of data generated by an online experiment. All curves are obtained by a single experimental run with the default hyper-parameters. The x-axis denotes the number of steps in the online dataset, and y-axis denotes the accuracy measured against the golden preference model. Overall, proxy preference models have higher prediction accuracy than others, though the accuracy decays over time as the online policy drifts away from the pairwise dataset distribution from which proxy preference models are trained.}
    \label{fig:classification-rm}
\end{figure*}

\subsection{Evidence for and against Hypothesis 3 (Better classification, better performance)}

There are two parts to this hypothesis: (1) Proxy preference model generally achieves higher classification accuracy than policy as classifier; (2) The performance gap between online and offline algorithms can be partly attributed to the difference of classification accuracy.

To briefly recall the background, notice that the loss function seeks to maximize the log ratio $\log \pi_\theta(y_w|x) / \pi_\theta(y_l|x)$ of the winning response $y_w$ over the losing response $y_l$. As such, the policy can be understood as a preference model (with point-wise parameterization) or a reward model. Indeed, given two responses $y_1,y_2$ and prompt $x$, the policy can produce the preference of $y_1$ over $y_2$ by the scalar prediction
\begin{align*}
    f_\theta(x, y_1,y_2) =  \log \frac{\pi_\theta(y_1|x)}{\pi_\theta(y_2|x)} - \log \frac{\pi_\text{sft}(y_1|x)}{\pi_\text{sft}(y_2|x)},
\end{align*}
where the log ratio of the SFT policy serves as a baseline. The sign of this prediction can be used as a classifier on whether $y_1$ is preferred to $y_2$ according to policy $\pi_\theta$.

\subsubsection{Proxy preference model is generally more accurate than policy (of the same size) as classifier}
Across various tasks, the proxy preference model can obtain about $70\%-90\%$ training accuracy while when using offline policies as classifiers of pairwise preference, the accuracy peaks at about $\sim 70\%$ during training (see Figure~\ref{fig:classification}). This means that when evaluated as classifiers, proxy preference models with the same size are generally more performant. 

In Figure~\ref{fig:classification-rm} we assess the out-of-distribution classification accuracy of  proxy preference model vs. policies. We compare three baselines: proxy preference models, online and offline policies trained for 4k steps. The dataset is the sequence of data points generated by the online experiment. Here, \emph{out-of-distribution} refers to the fact that as the training progresses, the sampling distribution drifts further away from the pairwise preference dataset from which the preference model and offline policies are trained on. Each data point in Figure~\ref{fig:classification-rm} shows the accuracy for the pairwise data generated near that training step.

As observed, the proxy preference model has generally significantly higher accuracy than both policy baselines. Noticeably, the accuracy of the proxy preference model also decreases over time, as the sampling becomes increasingly out-of-distribution for the proxy preference model. This can partially explain the over-optimization for online algorithms shown in Figure~\ref{fig:online-offline}.

\paragraph{Why are proxy preference models more accurate?}
While preference model makes a single scalar prediction, when using policy as classifiers one makes the prediction using likelihoods of both responses. One conjecture is that this is a much harder optimization problem and is less likely to generalize well. Also, preference models can take both responses into the context, making the prediction more expressive. A even more carefully controlled setting would be to apply a point-wise reward model rather than a preference model for comparison.

\begin{figure*}
    \centering
    \includegraphics[width=0.95\textwidth]{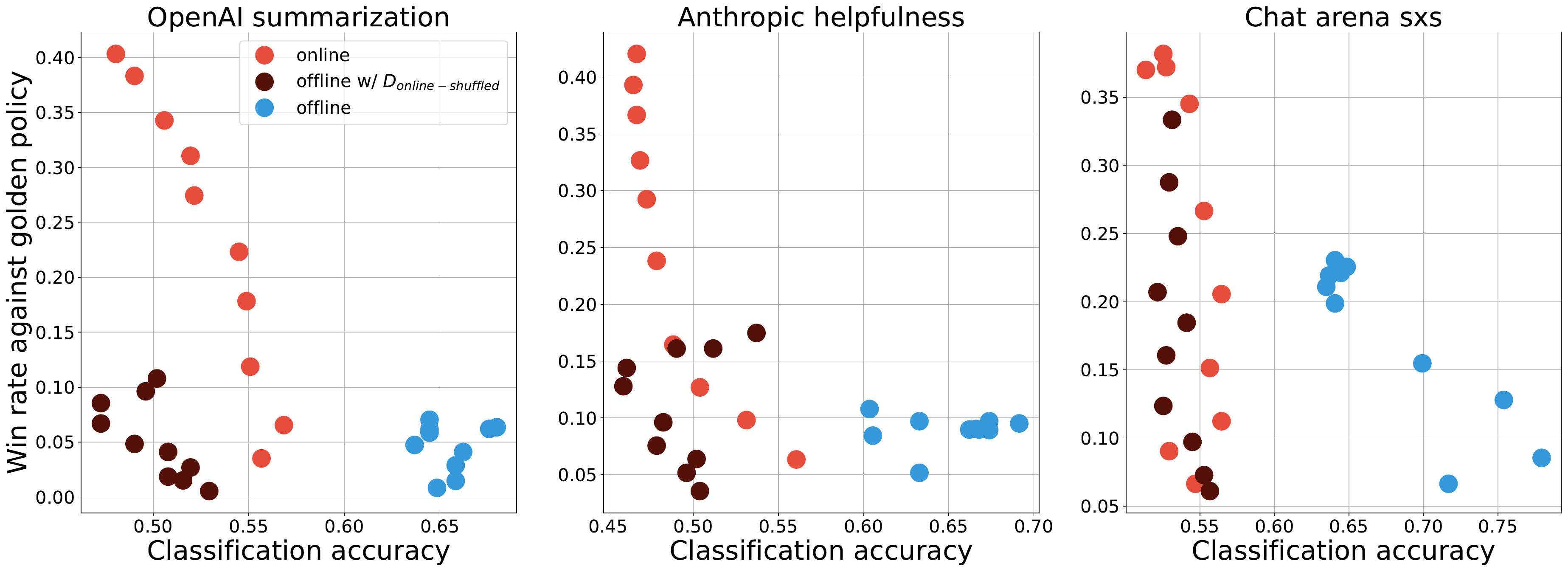}
    \caption{Policy performance as a function of the classification accuracy of the policy as classifier on pairwise preference dataset $\mathcal{D}_\text{golden}$. All results are obtained by a single online or offline experiment with the default hyper-parameter. Since offline experiments train policies to be better at classification on the preference dataset, the offline accuracy is generally higher. However, there is little positive correlation between the classification accuracy and policy win rate. This suggests that classification accuracy alone cannot explain the performance gap between online vs. offline experiments. All classification accuracy is measured against the golden preference model.}
    \label{fig:classification-winrate}
\end{figure*}

\subsubsection{Classification accuracy is not predictive of generative performance} Now that we have established the difference in the classification accuracy between proxy preference models and offline policies as classifiers, does this difference explain the performance gap between online and offline algorithms?

In Figure~\ref{fig:classification-winrate}, we plot the policy performance across various algorithms against their classification accuracy on the pairwise preference dataset $\mathcal{D}_\text{golden}$. Note that since offline policies are trained as classifiers on such datasets, they generally have higher prediction accuracy (compared to e.g., online policies). However, focusing just on the offline experiments, we find little statistical correlation between the classification accuracy and model performance. In other words, by just improving the policy's classification accuracy on the preference dataset (e.g., by changing loss function, model architecture or model size), it is unlikely to yield significant performance gains.

\subsubsection{Interplay between discriminative and generative performance} 

Now is a good moment to dive into the interplay between discriminative and generative abilities of policies trained from various algorithms. Examining the marginal correlation between classification accuracy and model performance across all algorithms, one might even conclude that improving classification accuracy hurts generative performance. Note that this is incorrect, as online algorihtms do not aim for improving discriminative capabilities on any fixed dataset.

\begin{figure*}
    \centering
    \includegraphics[width=0.95\textwidth]{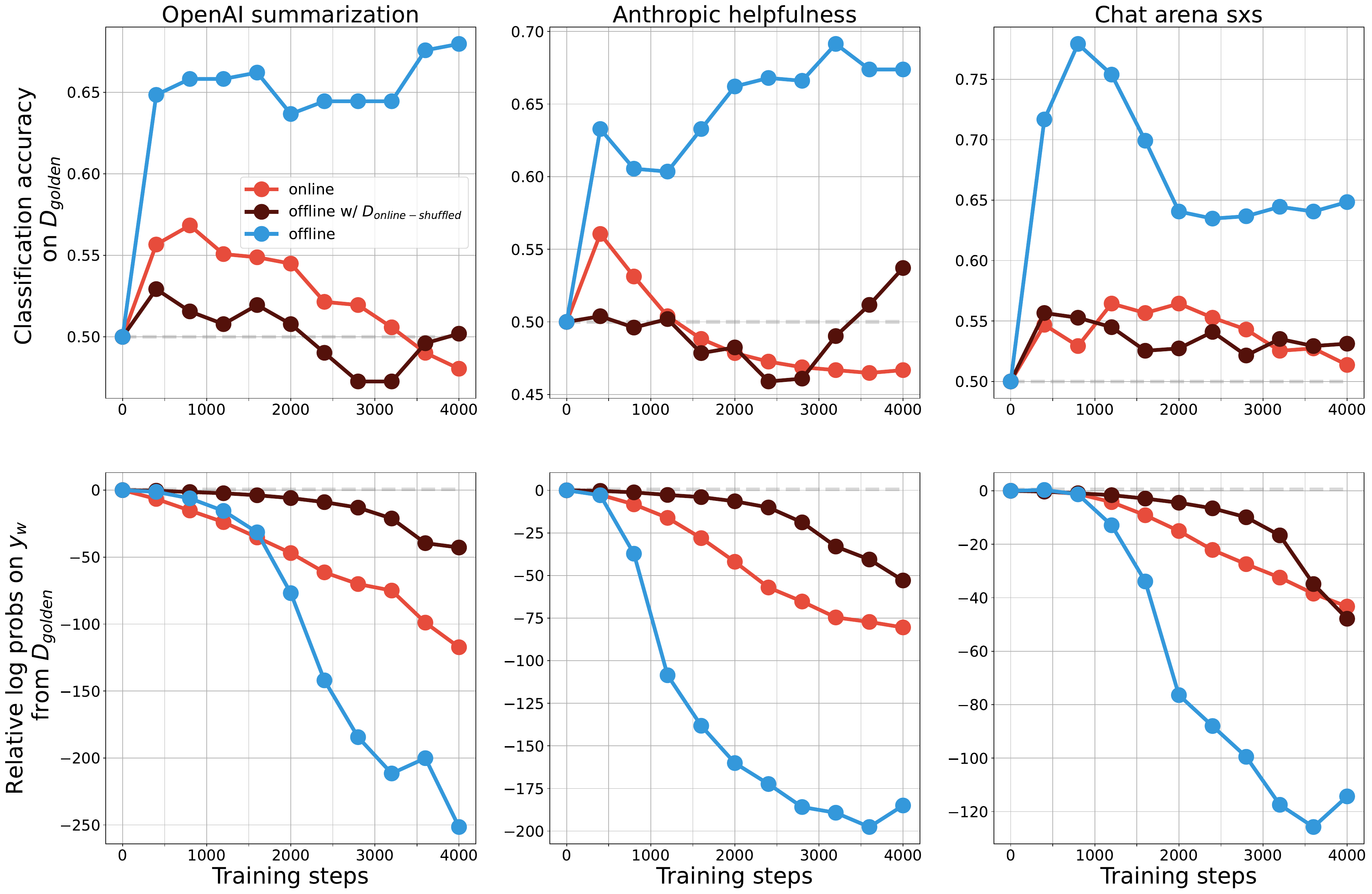}
    \caption{Top row: classification accuracy over the pairwise preference dataset when using the learned policies as preference classifiers. The accuracy is estimated by subsampling $256$ data points from the dataset. Offline experiments produce policies that can lead to $60\%-70\%$ classification accuracy, significantly higher than the other experimental settings (online). Bottom row: log probs of the winning responses from the pairwise dataset, relative to the SFT policy  $\mathbb{E}_{(x,y_w,y_l)\sim \mathcal{D}_\text{golden}}\left[\log \pi_\theta(y_w|x) - \log \pi_\text{sft}(y_w|x)\right]$. Over time, all policies place less weight on the winning responses and this trend is especially clear for the offline experiments. We note that this observation depends on various factors and especially dataset, and other work \citep{tajwar2024preference} reports cases where the decay in likelihood is less severe.}
    \label{fig:classification}
\end{figure*}

In Figure~\ref{fig:classification}, we study how the classification ability of policies on the preference dataset trained by various algorithms evolve over time. From the top row, we showcase the classification accuracy of learned policies over time, from online, offline and tandem experiments (offline with $\mathcal{D}_\text{online-shuffled}$, computed over the pariwise preference dataset.  The offline experiment behaves as expected, increasing the classification accuracy from random guess to $60\%\sim 70\%$ over time. This attests to the fact that the offline optimization process works \emph{as intended}, since the policy is becoming a better classifier over time. However, this is not the case for online experiments where the classification accuracy lingers about sub $50\%$. This means better generative ability (higher performance) does not necessarily translate into better classification accuracy, and vice-versa.

\paragraph{Why are online algorithm policies not trained to be good classifiers?} It might come as a surprise that online algorithm policies seem not to be trained as good classifiers, despite sharing the same loss function as offline algorithms. The key difference lies in the fact that unlike a static distribution, online algorithms apply a sampling distribution that changes over time. As a result, as soon as the policy makes an infinitesimal progress in the classification accuracy on the incoming batch, the sampling distribution moves away. The shifting learning target may pose a challenge for the model to reach high accuracy in classifying data sampled from a static dataset. When the model capacity is limited, this moving target might not be effective at teaching the model to classify any static distribution. However, during such a non-stationary learning process, the policy shifts probability masses towards more promising responses, and improves the generative capability over time.

\paragraph{Offline algorithms do not make winning responses more likely.} As an intriguing and perhaps surprising finding (Figure~\ref{fig:classification} bottom row), across all experiments, the likelihood of the winning responses from the pairwise preference dataset $\mathcal{D}_\text{golden}$ decrease over time. Similar observations were made in \citep{pal2024smaug,rafailov2024} under slightly different experimental settings. This implies a shift in sampling distributions for all experiments, and that means we cannot interpret contrastive losses as a form of contrastive SFT as motivated in Eqn~\eqref{eq:contrastive-sft}. Surprisingly, the decay in log probs is the most drastic for the offline algorithms. This suggests that in order to obtain high classification accuracy, a frequent solution of offline optimization is to set the likelihoods of winning responses low and losing responses even lower.

The fact that both likelihoods decreases for offline experiments, does not imply that offline algorithms would fail to learn, indeed our results thus far suggest otherwise (Figure~\ref{fig:online-offline}). This observation, however, implies that offline optimization does not shift more probability masses to winning responses sampled from the offline dataset, unlike online algorithms which shift probability masses towards increasingly better responses to improve generative performance. An explanation is that since online algorithms only observe likely responses under the current policy (thanks to on-policy sampling), it leads to a different optimization dynamics than offline learning. In sum, the generative ability of offline policy is improved through a much more indirect process than online policy.

\paragraph{Can policies accurately classify their own samples?} On a side note, an intriguing question is how accurately can policies classify their own samples.
Note that a theoretical optimal policy assigns probability masses to responses proportional to their quality, and should classify its own samples accurately too. In fact, this self-classification metric is closely related to policy improvement and we found that the online IPO loss (Eqn~\ref{eq:loss}) is closely related to this metric, yet the algorithm does not exactly optimize for it. In fact, throughout training, we found that policies cannot do statistically significantly better than random guess at their own samples (even for online experiments). This means that all the policies we have trained are likely all far from optimal. See Appendix~\ref{appendix:policy-as-classifier} for more discussions.

\subsection{Evidence against Hypothesis 4 (Non-contrastive loss function)}

As discussed before, one might question whether the under-performing behavior thus observed of offline algorithms are due to the contrastive losses, rather than offline samples. What if we consider loss functions with less of a contrastive formulation, do we still observe the performance gap? 

Tracing back to how contrastive losses were derived in the first place, the motivation was to increase the likelihood of winning responses and decrease the likelihood of losing responses. The contrastive losses offer a form of variance reduction, and hence more stable signals for improvements. To study the effect of offline vs. online sampling, and to remove the confounding factors due to the optimization \emph{pathologies} of contrastive losses suggested in Figure~\ref{fig:classification}, we consider the Best-of-2 (Bo2) loss 
\begin{align}
   \min_\theta \mathbb{E}_{x\sim p,(y_w,y_l)\sim \mu}\left[\log \pi_\theta(y_w|x)\right]\label{eq:bo2-loss}
\end{align}
which is a canonical non-contrastive loss and has been studied in the over-optimization setup in the one-step case \citep{gao2023scaling}.
For both online and offline instantiations, the algorithm effectively carries out SFT on the winning responses out of the sampled pair.

\begin{figure*}
    \centering
    \includegraphics[width=0.95\textwidth]{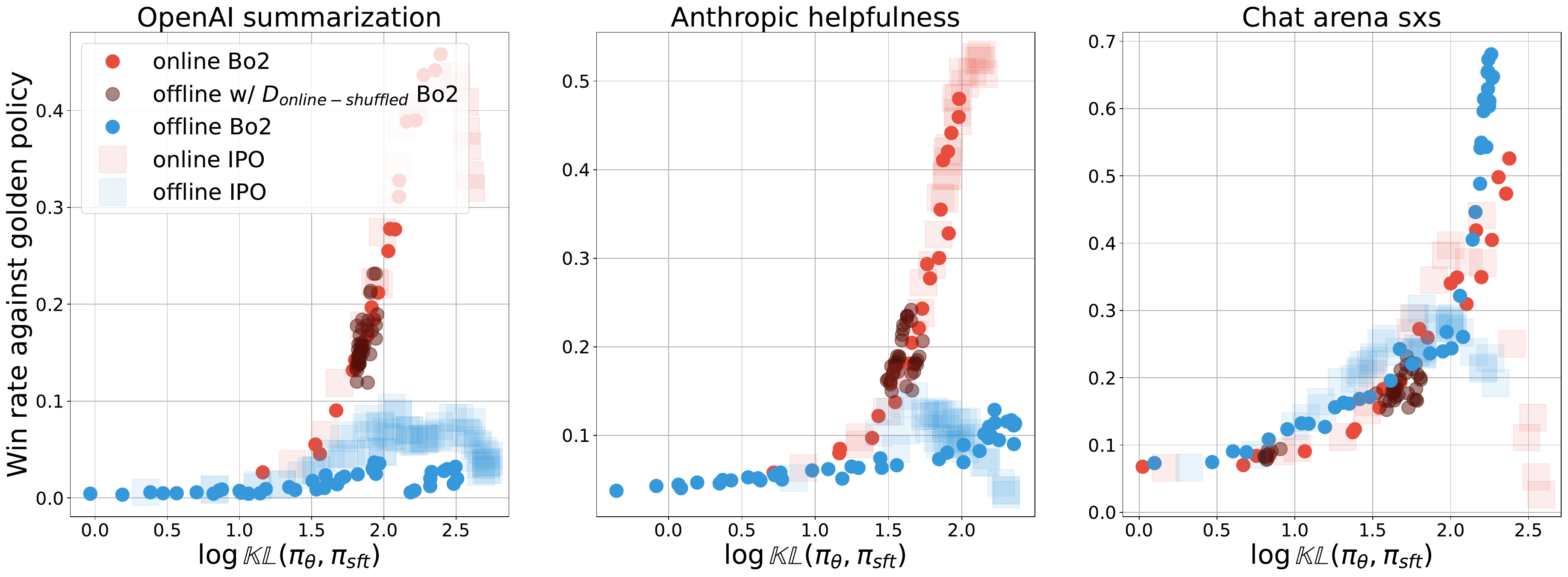}
    \caption{Trade-off between KL divergence vs. performance for the Best-of-2 (Bo2) loss function. The experimental setup is identical to Figure~\ref{fig:online-tandem-offline} except for the loss function. The performance gap persists between offline and online experiments, except for Chat arena sxs, where there seems to be little difference in the trade-off curves (or rather, offline can obtain slightly better trade-off when the budget is high). The offline experiments with online generated dataset $\mathcal{D}_\text{online-shuffled}$ seems to be on par with the trade-off exhibited by the online experiments, indicating that the data coverage hypothesis might play a bigger role here.}
    \label{fig:bo2}
\end{figure*}

In Figure~\ref{fig:bo2}, we show the comparison between online and offline variants based on Bo2, akin to how Figure~\ref{fig:online-tandem-offline} is to the IPO loss. For ease of comparison, we also graph IPO results in the background.  We make a few observations:
\begin{itemize}
    \item \textbf{Online vs. offline gap for Bo2}. The online vs. offline comparison is similar as before, except for the Chat arena sxs task where the offline algorithm seems to perform much better and the rate of improvement even exceeds that of the online experiment. This result might be suggestive of the nature of the dataset, for chat arena sxs, it might suffice to SFT on the winning response. The performance gap in other two datasets still showcases the difference between online and offline algorithms as illustrated before.
    \item \textbf{Data coverage hypothesis might offer a better explanation now}. We perform offline Bo2 with $\mathcal{D}_\text{online-shuffled}$ similar as before. The trade-off curve matches quite well with the online algorithm, within the set of KL divergence the optimization induces. This implies that for Bo2, the performance difference between online and offline is better captured by the data coverage hypothesis.
\end{itemize}

Even though the data coverage hypothesis might provide a better explanation to the online vs. offline performance gap for Bo2, once again it is important to note that arriving at the dataset $\mathcal{D}_\text{online-shuffled}$ in the first place is infeasible. This dataset consists of responses generated by a sequence of evolving policies and is thus not available in practical situations.

\subsection{Evidence against Hypothesis 5 (Scaling policy is all you need)}

\begin{figure*}
    \centering
    \includegraphics[width=0.95\textwidth]{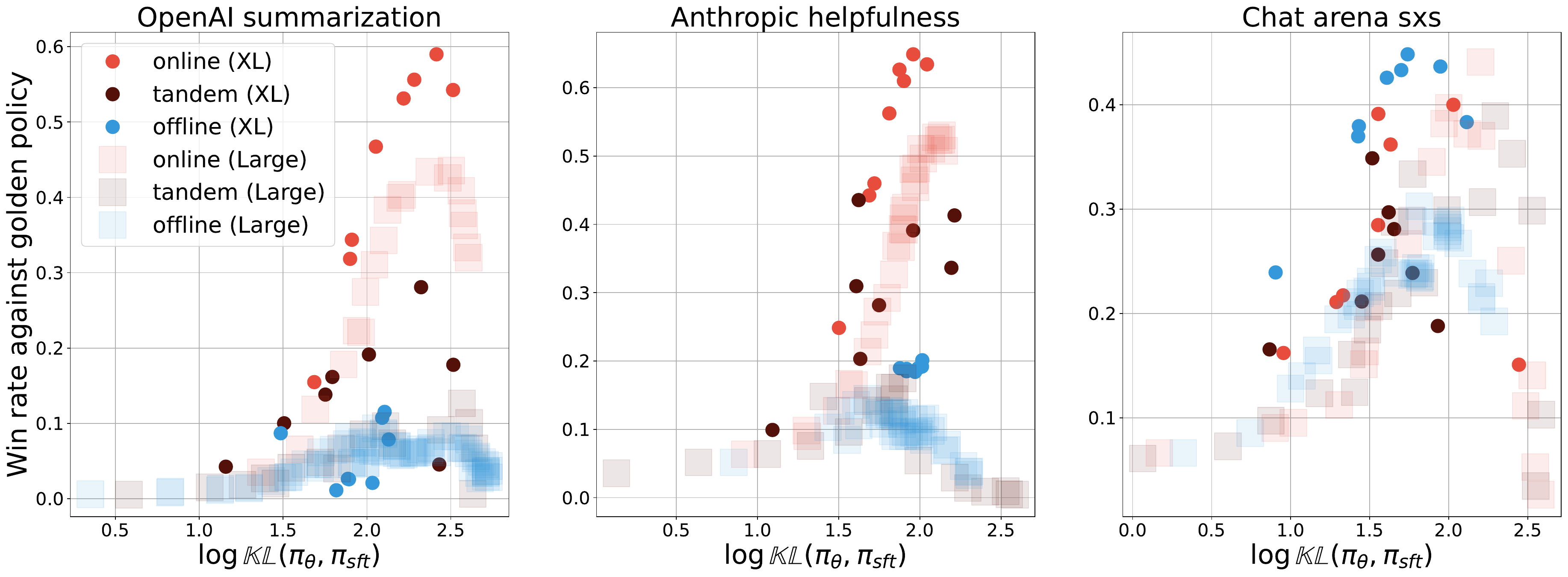}
    \caption{Trade-off of KL divergence vs. performance across two policy network sizes (XL and Large). We use \emph{tandem} to denote the offline experiments with shuffled online dataset $\mathcal{D}_{\text{online-shuffled}}$. As the policy size scales up, the peak performance becomes higher though obtained at a similar KL divergence budget as the smaller sized experiments. The improvement from Large to XL is potentially less clear for the Chat arena sxs tsak, where online and offline experiments now perform similarly.}
    \label{fig:scaling}
\end{figure*}

Finally, we ask if scaling solve some of the problems with offline optimization? We carry out online and offline experiments with XL (3B parameters) and XXL (11B parameters) models as policy networks, and use the same proxy preference model. For networks of various sizes, we apply a similar recipe for SFT and RLHF. We lower the batch size and training steps accordingly due to compute constraints. All evaluations are measured against the same golden policy using the golden preference model.

In Figure~\ref{fig:scaling}, we juxtapose the KL divergence vs. performance trade-off plots between XL and Large models. We denote as \emph{tandem} experiment the offline experiments with shuffled online generated dataset $\mathcal{D}_{\text{online-shuffled}}$ for convenience. We make a few observations from the results.
\begin{itemize}
    \item \textbf{Over-optimization and peak performance}. Across both model sizes we observe the over-optimization effect of various experiments. This corroborates that the bottleneck is the proxy preference model, and scaling policy sizes will not make a fundamental impact \citep{gao2023scaling}. The peak performance seems to improve as the policy size increases.
    \item \textbf{Peak performance is obtained at similar KL budget across model sizes}. For policies of both sizes, we observe that the peak performance is obtained at a similar KL divergence, compatible with results in \citep{gao2023scaling}. We note that the KL divergence is not comparable between the two policy sizes since they are with respect to different SFT policies.
\end{itemize}

\begin{figure*}
    \centering
    \includegraphics[width=0.95\textwidth]{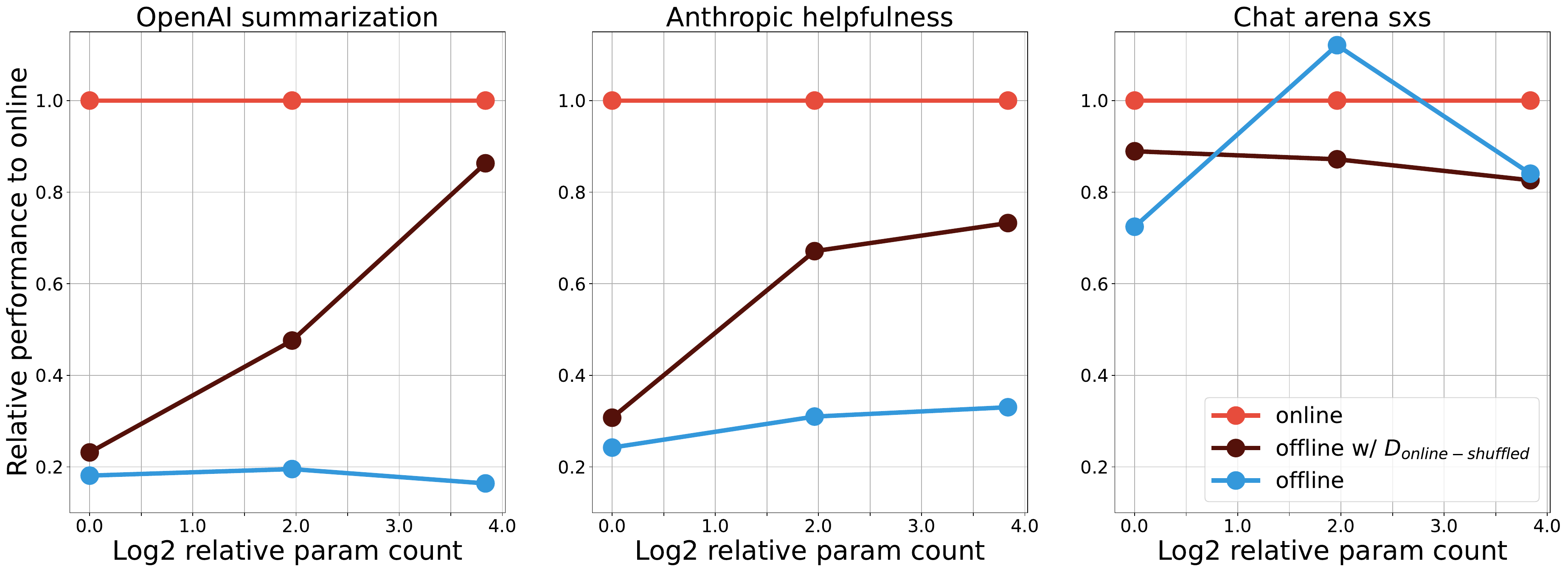}
    \caption{Best possible performance obtained from different algorithmic variants (online vs. offline vs. tandem) across different policy network sizes. For each network size, the best performance is obtained by taking the top results across all experiments and hyper-parameters for that network size. We normalize the performance with respect to the online policy, so that the online policy obtains the best possible performance of $1$. As the model size increases, the best possible gap between online and offline with $\mathcal{D}_\text{online-shuffled}$ decreases for two out of three tasks, indicating a bigger role by the data coverage hypothesis. However, the performance gap between online and offline experiments closes at a slower rate.}
    \label{fig:best-scaling}
\end{figure*}

In Figure~\ref{fig:best-scaling} we show how the \emph{best possible performance} scales as a function of policy sizes. Here, the best possible performance is obtained as the $90\%$ quantile of across all experiments under a particular method (online vs. offline) and with a fixed model size. For every policy size, we normalize all win rate with respect to the best possible online policy at for that size. The results are less clear for the Chat arena sxs task, while for the other two tasks, scaling policy networks by 16x (note the log scale on the plot) helps significantly bridge the gap between online and offline with online generated dataset $\mathcal{D}_{\text{online-shuffled}}$. This indicates that as the mode size increases, it is more likely that the online vs. offline performance is explained by the data coverage hypothesis.

The gaps between online and offline experiments also decrease but at a much lower rate. It is also not clear whether the gap will plateau at certain point, indicating a potentially divergent gap that cannot be bridged by just scaling the policy network.

\paragraph{Scaling up the proxy preference model.} If we scale up the proxy preference model, the peak performance for the online algorithms will expect to improve \citep{gao2023scaling}, further enlarging the online vs. offline gap. However, it is also important to note that the compute overhead of online vs. offline also increases proportionally.

\section{Making the dataset more on-policy improves offline learning} \label{sec:improve}

So far we have invalidated most of the hypotheses which might interpret the performance gap between online and offline algorithms. We now conduct another ablation study focusing on the curation of data for offline algorithms, to examine what property of the offline dataset leads to performance improvement.

The ablation datasets are constructed with the following properties which directly influence the quality of the dataset, most notably: (1) proxmity to SFT, which makes the dataset more on-policy during initial stage of training; (2) differentiations between pairwise responses, which arguably provide more steep learning signals to the contrastive learning loss; (3) absolute quality of responses, which measures the absolute quality of individual responses in the dataset. Note that in general these properties are not independent from one another, but we seek to identify marginal properties that lead to more consistent improvements. In practice, it is important to be mindful of their interactions and combinations.

\begin{figure*}
    \centering
    \includegraphics[width=0.96\textwidth]{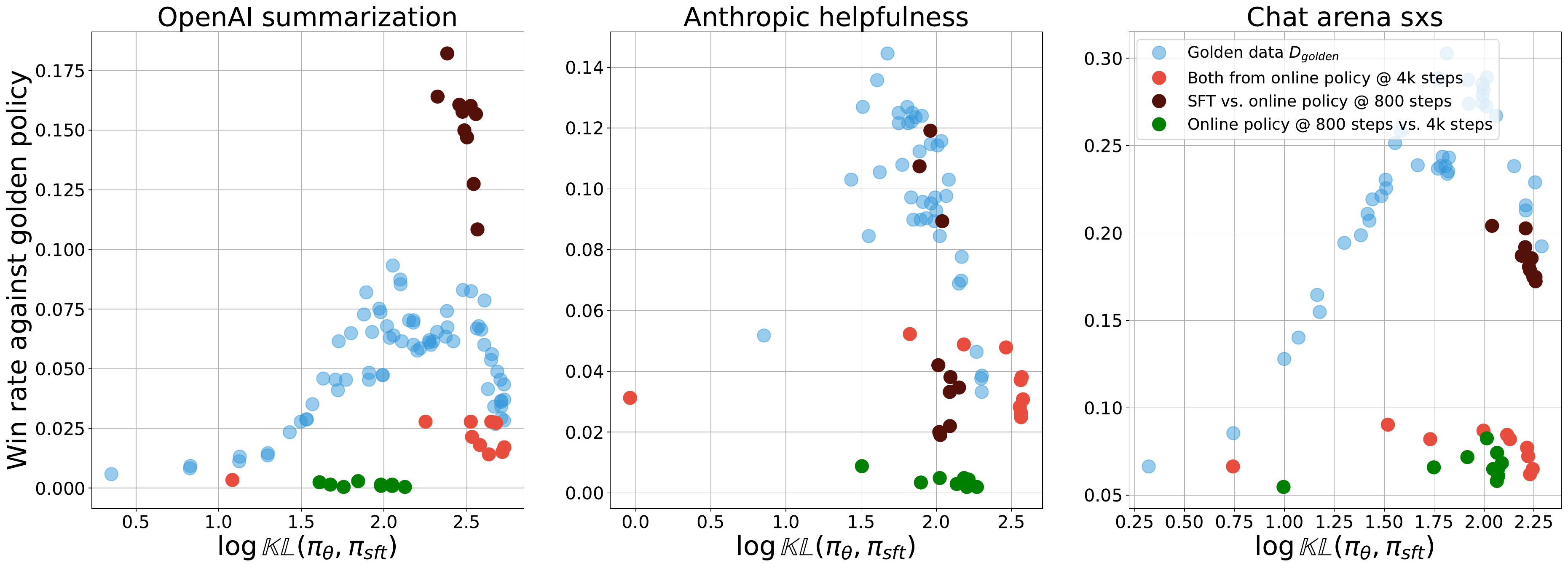}
    \caption{Dataset ablation for offline algorithms. We vary the offline datasets and assess how the policy performance changes as a result. Ablation datasets differ in how two sides of the responses are generated, we consider four alternatives: (1) Golden preference data $\mathcal{D}_\text{golden}$ which is the baseline; (2) Both sides are generated by online IPO policy trained for 4k steps $\mathcal{D}_\text{4k vs. 4k}$; (3) SFT vs. online IPO policy trained for 800 steps $\mathcal{D}_\text{sft vs. 800}$; (4) Online IPO policy trained for 800 steps vs. 4k steps $\mathcal{D}_\text{800 vs. 4k}$.}
    \label{fig:offline-dataset}
\end{figure*}

We consider two additional offline datasets for the ablation, differing in how two sides of the responses are generated. All prompts are sampled from the same dataset $\mathcal{D}_\text{golden}$ and at the end, paired responses are both scored by the golden preference model.
\begin{itemize}
    \item $\mathcal{D}_\text{sft vs. 800}$: Two sides of the responses are generated by the SFT policy $\pi_\text{sft}$ and the online algorithms' learned policy at 800 step.
    \item $\mathcal{D}_\text{800 vs. 4k}$: Two sides of the responses are generated by the online algorithms' learned policy at 800 step and 4k step.
    \item $\mathcal{D}_\text{4k vs. 4k}$: Two sides of the responses are both generated by the online policy at 4k step. This borrows the results from evidence against Hypothesis 2 (Sub-optimal offline dataset).
\end{itemize}
As a baseline, we compare the above datasets against $\mathcal{D}_\text{golden}$. Note that since our SFT policy is trained on both responses from the pairwise dataset, using the above notations, we can reinterpret the dataset as $\mathcal{D}_\text{sft vs. sft}$ (given that we can approximate the preference dataset $\mathcal{D}_\text{golden}$ by sampling both sides from SFT). By comparing the trade-off performance obtained via two sets of different datasets, we can verify the importance of property (1)-(3) introduced above. For example, comparing $\mathcal{D}_\text{800 vs. 4k}$ vs. $\mathcal{D}_\text{sft vs. 800}$ will provide evidence to the importance of the dataset's proximity to SFT, since both datasets consist of responses with big contrast.

\paragraph{Differentations between responses \emph{alone} do not seem effective at improving}
Figure~\ref{fig:offline-dataset} shows the KL divergence vs. performance trade-offs obtained by offline experiments on various offline datasets. We see that simply enlarging the differentiations between two sides of the paired responses do not improve the offline trade-off performance (compare $\mathcal{D}_{\text{800 vs. 4k}}$ vs. $\mathcal{D}_{\text{4k vs. 4k}}$ and compare $\mathcal{D}_{\text{800 vs. 4k}}$ vs. $\mathcal{D}_\text{golden}\approx \mathcal{D}_{\text{sft vs. sft}}$). An explanation is that when too off-policy, the learning process may fail to benefit from the contrastive information in the data.

\paragraph{Proximity to SFT seems to be most effective.}
Meanwhile, sampling one at least one side of the response from the SFT policy seems critical to improving the general trade-off (compare $\mathcal{D}_{\text{sft vs. 800}}$ vs. $\mathcal{D}_{\text{800 vs. 4k}}$). An interpretation of the above comparison is that by sampling at least one side of the responses from SFT, the dataset distribution becomes more on-policy during the initial stage of offline learning, and can at least make a solid one-step improvement against the SFT policy as with the initial stage of online learning. The offline dataset $\mathcal{D}_\text{golden}\approx \mathcal{D}_\text{sft vs. sft}$ also seems hard to beat, since the dataset distribution is close to the SFT policy to by construction.

Comparing $\mathcal{D}_{\text{sft vs. 800}}$ vs. $\mathcal{D}_\text{golden}\approx \mathcal{D}_\text{sft vs. sft}$, while both datasets are by design more on-policy against the SFT policy, it is not clear which one is dominantly better. Their difference lies in the differentiations between two sides of the responses, and the impact seems dataset dependent.

\section{Related work} \label{sec:related}

We discuss the connection between our results in relation to prior work on RLHF and offline vs. online RL algorithms in general.

\paragraph{Over-optimization in RLHF.} \citet{gao2023scaling} introduced the scaling law for reward model over-optimization in RLHF, where they studied the impact of reward models sizes. Our work is complementary in several ways: we provide the first comparative study between online and offline RLHF algorithms under a synthetic setup. As part of our technical contribution, we also provide the over-optimization trade-off for pairwise preference model rather than point-wise reward model.

While \citet{gao2023scaling} focused on the scaling of reward models, we provide results on the scaling of policy networks and observe how the online vs. offline gap persists as the network size scales up.

\paragraph{Contrastive losses for RLHF.} Some of the performance gap between online and offline algorithms  (Figure~\ref{fig:online-offline}) is attributable to the \emph{pathology} of contrastive losses, which are the dominantly applied in RLHF practices, though we have also identified similar performance gap in the case of non-contrastive losses (Figure~\ref{eq:bo2-loss}). Introducing a SFT based loss (such as Bo2) would mitigate such drawbacks on certain datasets (Figure~\ref{fig:bo2}), such methods do come with their own shortcomings (e.g., when both responses are of low quality and it is undesirable to apply a SFT loss on either response). It is an empirical question what is the ideal combination of contrastive and non-contrastive losses, which we leave open to the audience. See e.g., \citet{zhao2023slic,ethayarajh2024kto,pal2024smaug} for such example combinations.

\paragraph{Online vs. offline RL gap.} Our work is heavily inspired by the identification of \emph{tandem effect} in RL \citep{ostrovski2021difficulty}, where they side by side compared two RL agents, both applying the same loss function and observing the same stream of data. The data stream is actively generated by one of the agents, while the other agent could only learn passively. It is interesting to see that the challenge of offline learning is still prominent in RLHF, despite the various algorithmic differences from regular RL. For example, IPO is adopts a bandit formulation and does not bootstrap like Q-learning \citep{mnih2015human}. These changes were assumed to reduce the online vs. offline discrepancy but apparently the gap is still significant.

Concurrent to this work, \citet{xu2024dpo} studied the performance discrepancy between DPO and PPO for model alignment, focusing on implementation improvements to PPO. \citet{tajwar2024preference} studied the conditions under which techniques such as on-policy sampling and contrastive losses improve offline algorithms. They also ablate on when online algorithms do not obtain a clear gain over offline algorithms, such as when the reward model peak is within the offline distribution, which we do not investigate.  While all such work reach similar observations that on-policy sampling is important to alignment performance, we focus on understanding the causes of the discrepancy. 

\paragraph{Boundary between online and offline.} In practice, the boundary between online and offline is often quite blurred, since the offline dataset can evolve over time and making the entire training process effectively more on-policy. This is the case for a series of recent work 
\citep{zelikman2022star, gulcehre2023reinforced,singh2023beyond} though they focus on domains where golden rewards are available, and are less concerned with studying the impact of over-optimization.

\paragraph{Theoretical results on RLHF.} 
While theoretical research has proved fruitful in deriving variants of offline alignment algorithms \citep{rafailov2023direct,azar2023general,munos2023nash} and sought to elucidate the nature of empirical observations through a theoretical lens \citep{rafailov2024}, our observations suggest a retrospection of the assumptions made in theoretical derivations. One concrete example is the common yet implicit assumption that the dataset distribution has full coverage over the space of responses. As a result, empirical algorithms might work in ways that are not fully captured by theory.

\section{Conclusion}
Our work sheds light on the importance of on-policy sampling for LLM alignment, and unveils the challenges for offline alignment.
Our investigation started by measuring the performance gap between online and offline algorithms. We hypothesized about the cause of the gap from the perspectives of data, optimization procedure, loss function and scaling properties, and carefully designed a set of, we carefully designed a set of experiments and provided compelling evidences that largely invalidates many intuitively sensible hypotheses. In contrast to many aforementioned hypotheses, our additional data ablation study demonstrated that on-policy data generation is key to the improvement of policy learning efficiency.

Overall, our experimental design and empirical findings implies the fundamental necessity of on-policy learning for AI alignment. A natural question which follows is whether these results mean that offline algorithms are bound to under-perform. Our response to this question is also negative. As discussed above, the dichotomy of online vs. offline is often inaccurate in practice, since an offline algorithm with a repeatedly updated data stream is effectively an online algorithm. As a result, offline learning can be made less likely to suffer from the shortcomings identified in this work, by being more careful with the data generation process in general.

Our results also point to a few future directions. For example, given our evidence that online policy is better at generation while offline policy is better at discrimination, would it be possible to achieve a better trade-off by leveraging the best of both worlds? More ablation can also be conducted in a few other aspects of the problem too, such as how the performance gap depends on the complexity of the task, and on the quality of the base model. Our empirical results also suggest potential directions for deeper theoretical understanding of RLHF.

\paragraph{Acknowledgement.} We thank David Abel for his detailed and valuable feedback on an earlier draft, Yujia Li, Rahma Chaabouni and Bilal Piot for discussions of early ideas. Lastly, we thank the various engineering teams at Google DeepMind for building infrastructure that makes this work possible.

\paragraph{Contribution statement.} Yunhao initiated the project, ran experiments, carried out analysis and wrote the draft. Zeyu provided initial observations that inspired the work. Yunhao and Daniel carried out iterative developments of hypothesis based on results. Yunhao, Daniel and Zeyu analyzed key experimental results. Yunhao, Daniele and Eugene built the research code base. Zeyu, Yong, Will, Yuan and Bernardo contributed to additional hypotheses. Yunhao, Daniel and Yuan structured the presentation of the results. Others participated in project discussions and minor edits to the paper.

\newpage

\bibliographystyle{plainnat}
\bibliography{main}

\newpage

\appendix
\section{Details of training preference models}
\label{appendix:preference}

\subsection{Background}
Throughout, we apply preference models which take a tuple $(x,y_1,y_2)$ and generates a scalar prediction between $0$ and $1$: $r_\theta(x,y_1,y_2)\in[0,1]$. At training time, given a dataset $\mathcal{D}$ of pairwise preference $(x,y_w,y_l)$, the model is trained by maximum likelihood
\begin{align}
    \max_\theta \mathbb{E}_{(x,y_w,y_l)\sim\mathcal{D}}\left[\log r_\theta\left(x,y_w,y_l\right)\right].\label{eq:preference-loss}
\end{align}
The global minimizer to the above optimization problem is the ground truth preference probability $r^\ast(x,y_1,y_2)=p(y_1\succ y_2|x)$. The preference model can be understood as a more expressive version of point-wise reward model, which assigns a single scalar to each response $r(x,y)$ and usually makes a Bradley-Terry model assumption for training \citep{christiano2017deep}. The preference model is akin to a model for auto side-by-side comparison, as is commonly practiced in LLM evaluation (e.g., \citep{vertexsxs}).

Note that the loss Eqn~\eqref{eq:preference-loss} might induce a positional bias in the model, i.e., $r_\theta(x,y_1,y_2)$ might generally assign more preference to the first response than the second response.
To alleviate the positional bias, we explicitly randomize the ordering of the two responses during training.
\begin{align}
    \max_\theta \mathbb{E}_{(x,y_w,y_l)\sim\mathcal{D}}\left[\frac{1}{2}\log r_\theta\left(x,y_w,y_l\right) + \frac{1}{2} \log \left(1-r_\theta\left(x,y_l,y_w\right)\right)\right].\label{eq:randomized-preference-loss}
\end{align}

\subsection{Inference details}
Since the positional bias cannot be removed perfectly, we find that the online RLHF process can easily exploit such remaining bias if we just use $r_\theta(x,y_1,y_2)$ to determine the preference of $y_1$ over $y_2$. This can lead to unintended behavior. Hence at inference time (for both training and evaluation), given a tuple $(x,y_1,y_2)$, we determine the preference of $y_1$ over $y_2$ using the sign of the difference
\begin{align*}
    r_\theta(x,y_1,y_2) - r_\theta(x,y_2,y_1).
\end{align*}
The sign is used to denote whether $y_1$ is preferred over $y_2$.

\subsection{Training details}
Each dataset consists of a training set and a validation set. We train the preference model on the training set, and monitor the prediction accuracy on the validation set. In general, we terminate the training roughly when the validation accuracy plateaus but do not enforce a strict rule for termination.

For the golden preference model (XXL), we train on the initial pairwise dataset (preference provided by the dataset raters) with a batch size of $16$. For the proxy preference model (Large), we train on the pairwise labeled by the golden preference model with a batch size of $32$ to match the batch size for the offline policy optimization algorithms. Throughout, the learning rate is $10^{-4}$ and applies and linear warm-up for 1k steps. The training context length is 1024 and target length is 128.

\section{Details of alignment algorithms}
\label{appendix:algorithm}

The IPO loss \citep{azar2023general} was derived as an alternative to the DPO loss. While the two algorithms employ different loss functions, they are both contrastive by design and seek to increase the likelihood of the winning response over the losing response, relative to the SFT policy. This is achieved by minimizing the general loss function parameterized by a convex function $f$:
\begin{align*}
   \min_\theta \mathbb{E}_{x\sim p,(y_w,y_l)\sim\mu}\left[f\left(\beta\left(\log\frac{\pi_\theta(y_w|x)}{\pi_\text{sft}(y_w|x)} -\log\frac{\pi_\theta(y_l|x)}{\pi_\text{sft}(y_l|x)} \right)\right)\right].
\end{align*}
For IPO $f(z)=(z-0.5)^2$ and for DPO $f(z)=\log(1+\exp(-z))$. Despite differences in the choice of the convex function, in general these algorithms obtain similar KL-divergence vs. performance trade-off as shown in \citep{tang2024generalized}. Hence, we should expect conclusions drawn about IPO to be transferable to other loss functions to some extent.

For both online and offline experiments, we apply a batch size of $32$ for the Large model, $16$ for the XL model and $8$ for the XXL model.

\section{Details of supervised fine-tuning}
\label{appendix:sft}

Throughout all the experiments, we supervise finetune the models as a starting point of the RLHF step. Instead of collecting separate SFT datasets, we repurpose all the RLHF pairwise preference datasets into a single SFT dataset. Concretely, we pool together datasets from all four downstream tasks. From each prompt $x$, we treat both responses $y_w,y_l$ as SFT targets. For the Large model, the SFT applies a batch size of $128$ and learning rate of $3\cdot10^{-5}$. The SFT runs for 4k steps, which constitutes about $3-4$ epochs on the entire dataset. The training context length is 1024 and target length is 128.

For the XL and XXL model, we finetune on the same SFT dataset for 8000 steps, at a learning rate of $10^{-5}$. For XL, the batch size is 16 and for XXL the batch size is 8. As a result, the XL model trains for about $\sim 1$ epoch on the entire dataset, while XXL trains about half an epoch. The post-SFT alignment performance suggests that this amount of finetuning seems enough to improve the overall performance as the policy network scales up.

The main objective of the SFT stage, is to ensure that the starting point of the RLHF step already achieves a reasonable performance, and not too out-of-distribution from the pairwise preference dataset that the preference models are trained on. This is also an attempt to make our experimental setup closer to practice, where models are expected to generate reasonable responses before RLHF \citep{christiano2017deep,openai2023gpt}.

\section{Details of estimating KL divergence}
\label{appendix:kl}
We present details on estimating the KL divergence
$\mathbb{KL}\left(\pi_\theta,\pi_\text{sft}\right)=\mathbb{E}_{x\sim \rho,y\sim\pi_\theta(\cdot|x)}\left[\log \frac{\pi_\text{sft}(y|x)}{\pi_\theta(y|x)}\right]$ in an unbiased way. By definition, we can construct unbiased estimates by sampling. We first sample $256$ prompts $(x_i)_{i=1}^{256}$ from the training set, and then for each prompt $x_i$, sample a response $y_i\sim \pi_\theta(\cdot|x_i)$ from the learned policy $\pi_\theta$. In general, the response consists of $T_i$ tokens, which we write as 
$y_i = \left(y_{i,t}\right)_{0\leq t\leq T_i-1}$. For each partial sequence of tokens $y_i = \left(y_{i,t}\right)_{0\leq t\leq j}$ with a fixed time step $j$, we can calculate the full distribution over the next token under both $\pi_\theta$ and $\pi_\text{sft}$:
\begin{align*}
    \pi_\theta\left(\cdot\; | \; \left(y_{i,t}\right)_{0\leq t\leq j}\right) \ \text{and}\ \pi_\text{sft}\left(\cdot\; | \; \left(y_{i,t}\right)_{0\leq t\leq j}\right)
\end{align*}
both of which are $V$-way categorical distributions with $V$ being the vocabulary size. The KL divergence between these two categorical distributions can be calculated explicitly for each step $j$, and we construct the full unbiased KL divergence as
\begin{align*}
    \frac{1}{256}\sum_{i=1}^{256} \sum_{t=0}^{T_i-1} \mathbb{KL}\left(\pi_\theta\left(\cdot\; | \; \left(y_{i,t}\right)_{0\leq t\leq j}\right), \pi_\text{sft}\left(\cdot\; | \; \left(y_{i,t}\right)_{0\leq t\leq j}\right) \right).
\end{align*}

For the XL model, when estimating the KL divergence we subsample $32$ prompts rather than $256$ prompts to reduce the inference burden.

\section{Using a policy as a classifier}
\label{appendix:policy-as-classifier}

Given a prompt $x$ and two responses $y_1, y_2$, we can calculate the log likelihood under the learned policy $\pi_\theta$ for each combination, and take the log ratio. We carry out the same process with the SFT policy, which was used as a reference during training.
\begin{align*}
    f_\theta(x,y_1,y_2) = \log \frac{\pi_\theta(y_1|x)}{\pi_\theta(y_2|x)} - \frac{\pi_\text{sft}(y_1|x)}{\pi_\text{sft}(y_2|x)}.
\end{align*}
If $f_\theta(x,y_1,y_2) > 0$ we classify that $y_1$ is preferred to $y_2$ and vice versa. This derivation comes naturally from the derivation of contrastive policy optimization losses such as DPO \citep{rafailov2023direct}, IPO \citep{azar2023general} and SLiC \citep{zhao2023slic}, as implementations of a binary classification problem \citep{tang2024generalized}. The decision boundary is defined by the sign of the scalar prediction $f_\theta(x,y_1,y_2)$.

All the classification accuracy reported in this work are obtained by subsampling the original datasets. For the evaluation on the pairwise preference dataset (Figure~\ref{fig:classification}), we subsample 256 prompts and pairs of responses and measure the preidction accuracy against the golden preference model. For the evaluation on the online generated data (Figure~\ref{fig:classification-rm}), we subsample 64 prompts and pairs of responses near each one of the 10 evenly spaced checkpoints from the online experiments.

\subsection{Policies as classifiers for their own samples}

The self-classification accuracy metric is
\begin{align*}
    \mathbb{E}_{x\sim\rho,(y_w,y_l)\sim \pi_\theta}\left[ \mathbb{I}\left[f_\theta(x,y_1,y_2) > 0\right]\right]
\end{align*}
where throughout, the preference judgement is obtained from the golden preference model. Approximating the non-linear identity function by a linear form such as $\mathbb{I}[z>0]\approx z$, we arrive at a surrogate loss
\begin{align*}
    \mathbb{E}_{x\sim \rho,(y_w,y_l)\sim \pi_\theta}\left[\log \frac{\pi_\theta(y_w|x)}{\pi_\theta(y_l|x)} - \log \frac{\pi_\text{sft}(y_w|x)}{\pi_\text{sft}(y_l|x)}\right],
\end{align*}
which is bears close connections to the linear part of the online IPO loss. However importantly, note that to optimize the surrogate loss above, one needs to control the sampling distribution $(y_w,y_l)\sim \pi_\theta$, which is not optimized by the IPO loss. Indeed, the IPO loss only optimizes for the \emph{integrand} rather than the sampling distribution. Hence, we do not necessarily expect the online IPO policy to improve the self-classification accuracy metric over time (which empirically also does not improve).

\paragraph{Theoretical optimal policy is \emph{almost} an optimal classifier.} For offline algorithms, assume the offline data distribution is under a single policy $\mu$, then the theoretical optimal policy under the IPO loss is  \citep{azar2023general}
\begin{align*}
    \pi^\ast(y|x) \propto \pi_\text{sft}(y|x)\exp\left(\beta^{-1}p(y\succ \mu)\right)
\end{align*}
where $p(y\succ \mu|x)$ indicates the probability that response $y$ is preferred over responses generated under $\mu$, under the golden preference model. Then we have the scoring function under the optimal policy as
\begin{align*}
    f^\ast(x,y_1,y_2) = \frac{1}{\beta}\left(p(y_1\succ \mu) - p(y_2\succ \mu)\right).
\end{align*}
Now, the \emph{almost} part from the above statement comes from the fact that we need further assumption on the golden preference structure, to determine the accuracy of the optimal policy as a classifier. Assume the golden preference model follows a Bradly-Terry model (intuitively, this means all responses can be ordered), then  we have
\begin{align*}
    \text{sign}\left(f^\ast(x,y_1,y_2)\right) = \text{sign}\left(p(y_1 \succ - y_2) - \frac{1}{2}\right).
\end{align*}
In other words, the optimal policy can achieve $100\%$ classification accuracy. Similar results can be easily derived for DPO \citep{rafailov2023direct} and potentially generalized to other classes of loss functions \citep{tang2024generalized}.

\section{Tandem effect}
\label{appendix:tpu-tandem}

As described in the main paper, it is not difficult to see that running offline experiment on the online generated dataset $\mathcal{D}_{\text{online}}$ is \emph{mathematically} equivalent to the online experiment. Indeed, this is thanks to the identical policy initialization from the SFT policy, and the same set of hyper-parameters used during training.

During the initial investigation, we sought to verify this claim empirically. We made great efforts at ensuring that the data stream observed by the online experiments and the offline experiments are identical at token level. Simply making sure the exact match at token level across the entire training process, will ensure that the two learning paths stay close in a reasonably numerical sense. Interestingly, we found that even slight mismatch would cause drastic divergence in the learning curves. A few mismatching examples include (1) strings are the same but tokens differ; (2) missing 1-2 batches of tokens due to pre-emptions of large scale compute systems. These are consistent with observations from \citep{ostrovski2021difficulty}, which we term the \emph{offline tandem effect}.

After ensuring the data stream match, we observed that the offline experiment with $\mathcal{D}_\text{online}$ almost perfectly with online experiment, in terms of the training statistics (such as loss functions) up to $\sim 0.1\%$ throughout training. The discrepancy slowly increases over time, and towards the end of training can lead to visbly difference loss function. However, the learned policy has similar performance when measured against the golden policy baseline.

We conjecture that such slight discrepancy is caused by the inherent numerical randomness in the compute hardware, hinting at the inherent impossibility to numerically replicate the same experiment twice on modern hardware, an effect we term the \emph{hardware tandem effect}. Similar observations have been made in the RL case \citep{ostrovski2021difficulty}, where they observe even bigger performance difference. In our case, the discrepancy is to some degree mitigated by a regularization against the SFT policy.

\section{Additional results}\label{appendix:results}

\begin{figure*}
    \centering
    \includegraphics[width=0.45\textwidth]{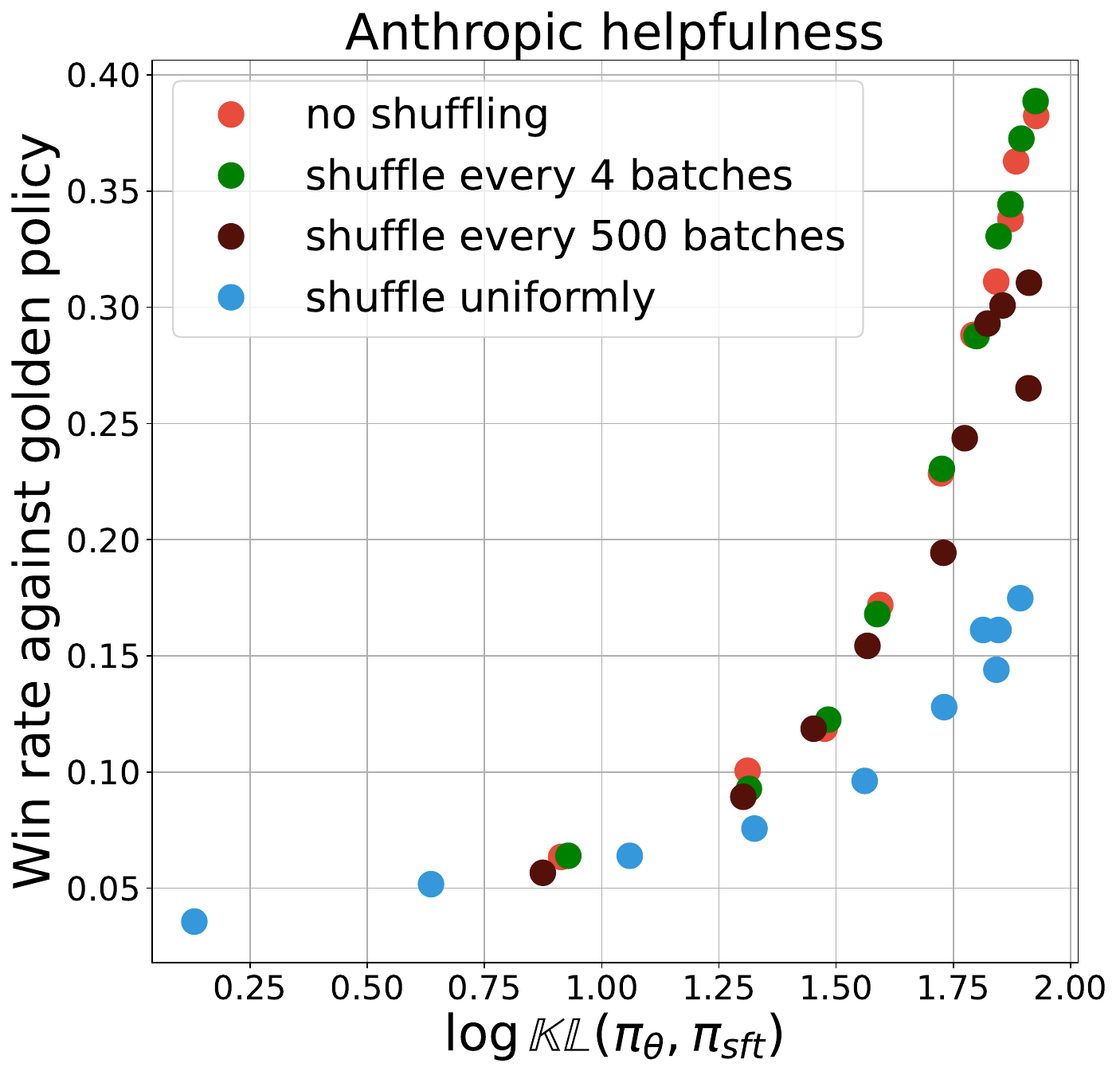}
    \caption{KL divergence vs. performance trade-off for offline experiments run on online generated dataset with different levels of shuffling. When the level of shuffling is small, the trade-off curve is similar to that of the online experiments. When the level of shuffling increases, given a fixed KL divergence budget, the performance decreases.}
    \label{fig:shuffle}
\end{figure*}

\paragraph{Performance as a function of level of shuffling.} Figure~\ref{fig:shuffle} shows the trade-off curves of different offline experiments trained on online generated dataset with various levels of shuffling. The performance is robust to a small amount of shuffling, that is, when the shuffling level is small, the performance does not change  significantly as a function of the KL divergence budget. However, when there is sufficient amount of shuffling, the performance starts to drop.

\begin{figure*}
    \centering
    \includegraphics[width=0.96\textwidth]{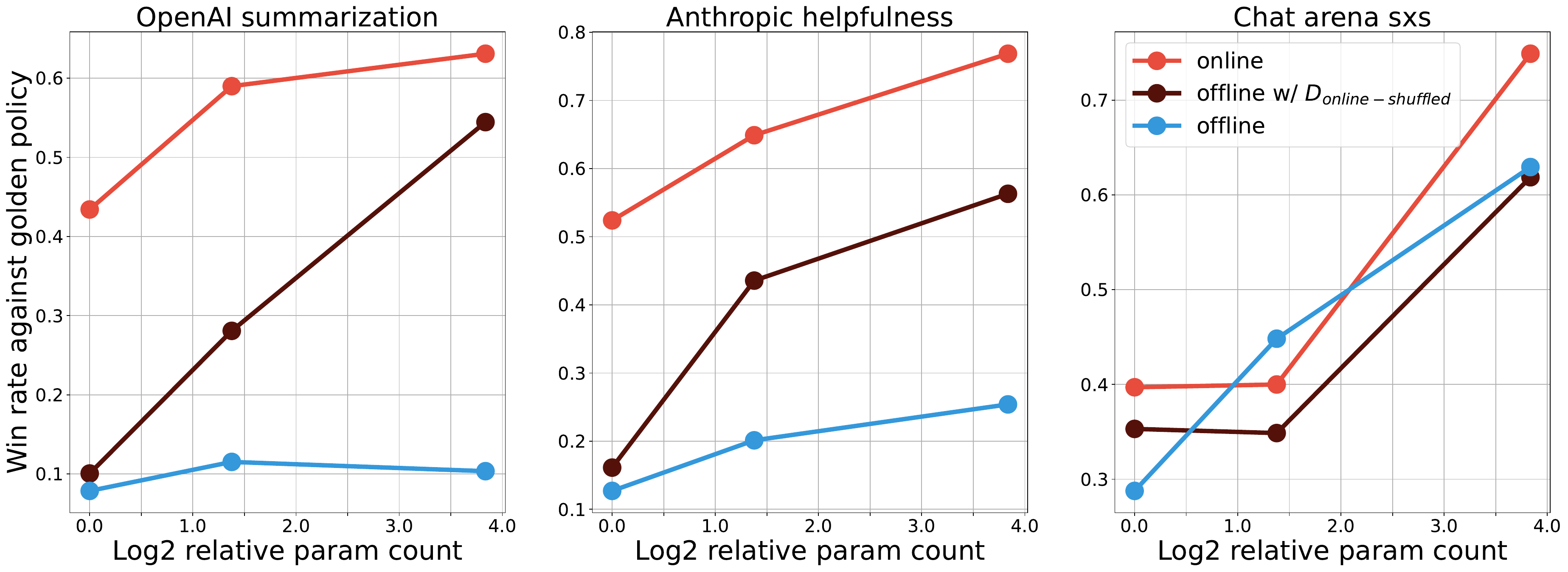}
    \caption{Best possible performance obtained from different algorithmic variants (online vs. offline vs. tandem) across different policy network sizes. The result is identical to that of Figure~\ref{fig:best-scaling} but here the performance is the absolute win rate against the golden policy.}
    \label{fig:best-scaling-abs}
\end{figure*}

\paragraph{Absolute win rate performance as a function of policy size.} Figure~\ref{fig:best-scaling-abs} shows the absolute win rate of policies obtained from various optimization algorithms, as a function of the policy network sizes. This result complements the relative performance results in Figure~\ref{fig:best-scaling}.

\section{An empirical case that current theory fails to predict} \label{appendix:theory-failures}

Let $\mu$ be the behavior policy or the dataset generation policy.
The offline IPO algorithm optimizes for the squared loss defined in Eqn~\eqref{eq:loss}. Now, \emph{assuming $\mu$ has full support over all possible responses}, the global minimizer to the offline IPO loss is
\begin{align*}
    \pi^*(y|x)\propto \pi_\text{sft}(y|x)\exp\left(\beta^{-1}p(y\succ \mu)\right).
\end{align*}
which alternatively, can be understood as optimizing the regularized objective \citep{azar2023general}
\begin{align*}
    \pi^\ast = \argmax_\theta \mathbb{E}_{x\sim\rho}\left[\mathbb{E}_{y\sim\pi_\theta(\cdot|x)}\left[\underbrace{p(y\succ \mu \;|\; x)}_{\text{win against $\mu$}} - \beta \mathbb{KL}\left(\pi_\theta(\cdot|x),\pi_\text{sft}(\cdot|x)\right)\right]\right],
\end{align*}
where the term $p(y\succ \mu|x)\coloneqq\mathbb{E}_{y'\sim \mu}\left[p(y\succ y')\right]$ denotes the win rate of response $y$ against responses generated from the behavior policy $\mu$, as judged by the golden preference model. In other words, we can understand the theoretically optimal solution as improving against the behavior policy $\mu$ in terms of the golden preference win rate.

\paragraph{Empirical observation \emph{conflicting} theoretic predctions.} When the dataset is collected using an highly-performant policy $\mu$ trained by online IPO for 4k steps, we observe that offline experiments barely take off in the performance as KL divergence increases. This means that the optimization process does not seem to improve upon the behavior policy $\mu$, at least not as predicted by theory.

We can interpret the above results as the breakdown of theory, in violation of the assumption that $\mu$ has full support. In practice, $\mu$ only has a small effective coverage over the space of responses.

\end{document}